# CORAL: expert-Curated medical Oncology Reports to Advance Language model inference

Madhumita Sushil[1,*], Vanessa E. Kennedy[2,#], Divneet Mandair[2,#], Brenda Y. Miao[1], Travis Zack[1,2,†], Atul J. Butte[1,2,3,4,†]

1. Bakar Computational Health Sciences Institute, University of California, San Francisco, USA

2. Helen Diller Family Comprehensive Cancer Center, University of California, San Francisco, USA

3. Center for Data-driven Insights and Innovation, University of California, Office of the President, Oakland, CA, USA

4. Department of Pediatrics, University of California, San Francisco, CA, 94158, USA

* Corresponding Author

Email ID: Madhumita.Sushil@ucsf.edu

Bakar Computational Health Sciences Institute, 490 Illinois Street, Cubicle 2215, 2nd Fl, North Tower, San Francisco, CA 94143

Telephone: +1 (415)-514-1971

# Equal contribution as the second author

† Equal contribution as the final author

# Abstract

**Background:** Both medical care and observational studies in oncology require a thorough understanding of a patient's disease progression and treatment history, often elaborately documented within clinical notes. As large language models (LLMs) are being considered for use within medical workflows, it becomes important to evaluate their potential in oncology. However, no current information representation schema fully encapsulates the diversity of oncology information within clinical notes, and no comprehensively annotated oncology notes exist publicly, thereby limiting a thorough evaluation.

**Methods**: We curated a new fine-grained, expert-labeled dataset of 40 de-identified breast and pancreatic cancer progress notes at University of California, San Francisco, and assessed the abilities of three recent LLMs (GPT-4, GPT-3.5-turbo, and FLAN-UL2) in *zero-shot* extraction of detailed oncological information from two narrative sections of clinical progress notes. Model performance was quantified with BLEU-4, ROUGE-1, and exact match (EM) F1-score metrics.

**Results:** Our team annotated 9028 entities, 9986 modifiers, and 5312 relationships. The GPT-4 model exhibited overall best performance, with an average BLEU score of 0.73, an average ROUGE score of 0.72, an average EM-F1-score of 0.51, and an average accuracy of 68% (expert manual evaluation on subset). Notably, GPT-4 was proficient in tumor characteristic and medication extraction, and demonstrated superior performance in advanced reasoning tasks of inferring symptoms due to cancer and considerations of future medications. Common errors included partial responses with missing information and hallucinations with note-specific information.

**Conclusions:** By developing a comprehensive schema and benchmark of oncology-specific information in oncology notes, we uncovered both the strengths and limitations of LLMs. Our evaluation showed variable zero-shot extraction capability of the GPT-3.5-turbo, GPT-4, and FLAN-UL2 models, and highlighted a need for further improvements, particularly in complex medical reasoning, before performing reliable information extraction for clinical research, complex population management, and documenting quality patient care.

## Introduction

Cancer care is complex, often involving multiple treatments across different institutions, with the majority of this complexity only being captured within the textual format of an oncologist's clinical note. Optimal clinical decision-making as well as research studies based on real-world data require a nuanced and detailed understanding of this complexity, naturally leading to wide-spread interest in oncology information extraction research[1]. Recently, large language models (LLMs) have shown impressive performance on several natural language processing (NLP) tasks in medicine, including obtaining high scores on United States Medical Licensing Examination (USMLE) questions[2,3], medical question answering[4], promising performance for medical consultation, diagnosis, and education[5], identifying key findings from synthetic radiology reports[6], biomedical evidence and medication extraction extraction[7], and breast cancer recommendations[8]. However, due to the lack of publicly available and comprehensively annotated oncology datasets, the analysis of these LLMs for information extraction and reasoning in real-world oncology data remains fragmented and understudied.

To date, prior studies on oncology information extraction have either focused on elements represented within ICD-O3 codes or cancer registries[9,10], or on a subset of cancer- or problem-specific information[11–15]. No existing information representation and annotation schema is adept enough to encompass comprehensive textual oncology information in a problem-, note type-, and disease-agnostic manner. Although similar frameworks are being created for tabular oncology data[16], efforts for textual data sources have been limited to pilot studies[17], surveys of oncology elements studied across different research contributions[18,19], and domain-specific schemas[20]. In this research, we aim to develop an expert-labeled oncology note dataset to enable evaluation of LLMs in extracting clinically meaningful, complex concepts and relations by: (a) developing a schema and guidelines for comprehensively representing and annotating textual oncology information, (b) creating a dataset of 40 oncology progress notes labeled according to this schema, and (c) benchmarking the baseline performance of the recent LLMs for zero-shot extraction of oncology information. Sample workflow is demonstrated in **Figure S1 (**Supplementary Materials**)**.

# Methods

## Oncology-specific information representation schema

To holistically represent oncology information within clinical notes, we developed a detailed schema based on a hierarchical, conceptual structure of oncology information (also called frame semantics)[17,18,20], agnostic to cancer and note types under consideration. It comprised of the following broad concepts: patient characteristics, temporal information, location-related information, test-related information, test results-related information, tumor-related information, treatment-related information, procedure-related information, clinical trial, and disease state. Broad concepts further encompassed expert-determined fine-grained concepts, for example, radiology test, genomic test, and diagnostic lab test were represented within the "tumor test" category. The schema was implemented through three annotation modalities: a) entities or phrases of specific types, b) attributes or modifiers of entities, and c) relations between entity pairs. These relations could either be (i) descriptive, for example relating a biomarker name to its results, (ii) temporal, for example indicating when was a test conducted, or (iii) advanced, for example, relating a treatment to adverse events caused due to it. Together, the schema comprised of 59 unique entities, 23 attributes, and 26 relations (Supplementary Materials, **Table S1**). — The concepts and relationships annotated within this new schema incorporate nuanced details like symptom history attributed to the diagnosed cancer, clinical trials considered for patient enrollment, genomic findings, reasons for switching treatments, and detailed social history of the patient not otherwise represented in cancer registries and structured medical record, and are significantly more inclusive and specific than those extracted by existing clinical NLP pipelines like cTakes[21] and DeepPhe[22]. An openly-available clinical NLP model, Stanza[23,24], was used to pre-highlight the mentions of problems, treatments, and tests within text to aid the annotators. Elaborate annotation guidelines are provided in the supplementary materials, and the annotation schema in the format of an open-source annotation software, BRAT[1], is shared along with the source code at https://github.com/MadhumitaSushil/OncLLMExtraction.

## Data

We collected 20 breast cancer and 20 pancreatic cancer patients from the University of California, San Francisco (UCSF) Information Commons, containing patient data between 2012–2022, de-identified as previously described[25]. All dates within notes were shifted by a random, patient-level offset to maintain anonymity[25]. Only patients with corresponding tabular staging data, documented disease progression, and an associated medical oncology note were considered for document sampling. Some gene symbols, clinical trial names and cancer stages were inappropriately redacted in our automated handling, and these were manually added back to the clinical notes under the UCSF IRB #18-25163 and #21-35084. These two diseases were chosen for their dissimilarity — while breast cancer is frequently curable, heavily reliant on biomarker and genetic testing and treatment plans integrating radiation, surgical, and medical oncology, pancreatic cancer has high mortality rates, and highly toxic traditional chemotherapy regimens. All narrative sections except direct copy-forwards of radiology and pathology reports were annotated using the knowledge schema described above by one of two oncology fellows and/or a medicine student. This resulted in a final corpus of 40 expert-annotated clinical notes, which is available freely through the controlled-access repository *PhysioNet*[2] after signing a data-use agreement, with additional 100 notes each for breast and pancreatic cancer automatically labeled by GPT-4 using the same prompts as the benchmarking tests in this study.

---

[1] https://brat.nlplab.org/

## Zero-shot LLM extraction baseline

To establish the baseline capability of LLMs in extracting detailed oncological history, we evaluated three recent LLMs without any task-specific training (i.e. "zero-shot" extraction): the GPT-4 model[26], the GPT-3.5-turbo model (base model for the ChatGPT interface[27]), and the openly-available foundation model FLAN-UL2[28] on the following tasks derived from two narrative sections of clinical progress notes for breast and pancreatic cancer: *History of Present Illness* (HPI) and *Assessment and Plan* (A&P):

1) **Symptom presentation:** Identify all symptoms experienced by the patient, symptoms present at the time of cancer diagnosis, and symptoms experienced due to the diagnosed cancer, all further related to the datetime of their occurrence.

2) **Radiology tests:** List radiology tests conducted for the patient paired with their datetime, site of the test, medical indication for the test, and the test result.

3) **Genomic tests:** List genetic and genomic tests conducted for the patient paired with the corresponding datetime and the test result.

4) **First cancer diagnosis date:** Infer the datetime for the first diagnosis of cancer for the patient.

---

[2] Users will need to create an account on https://physionet.org/ and sign a data use agreement before downloading the dataset. Upon manuscript publication, the dataset would be searchable using the manuscript title here: https://physionet.org/content/. A valid CITI training certificate will need to be uploaded before signing the DUA.

5) **Tumor characteristics:** Extract tumor characteristics in the following groups: biomarkers, histology, stage (TNM and numeric), grade, and metastasis (along with the site of metastasis and the procedure that diagnosed metastasis), all paired with their datetime.

6) **Administered procedures:** Identify all interventional procedures conducted for the patient paired with their datetime, site, medical indication, and outcome.

7) **Prescribed medications:** List medications prescribed to the patient, linked to the beginning datetime, end datetime, reason for prescription, continuity status (continuing, finished, or discontinued early), and any hypothetical or confirmed adverse events attributed to the medication.

8) **Future medications:** Infer medications that are either planned for administration or discussed as a potential option, paired with their consideration (planned or hypothetical) and potential adverse events discussed in the note.

We used the GPT models via the HIPAA-compliant Microsoft Azure OpenAI studio and application programming interface, so that no data was permanently transferred to or stored by Microsoft or OpenAI similar to previously described[29]. Separately, we implemented the openly-available FLAN-UL2 model on the internal computing environment. Model inputs were provided in the format *{system role description} {note section text, prompt}*. GPT model settings and task-specific prompts are provided in the Supplementary Materials. Examples of structured output format were provided with prompts to enable automated evaluation.

## Evaluation

An automatic quantitative evaluation was performed by comparing the manually annotated, related entity pairs to the corresponding model output. Since LLMs generate free-text outputs to represent entity mentions and their relations, model performance was quantified using two evaluation metrics for comparing pairs of text — BLEU-4 with smoothing[30] and ROUGE-1[31]. BLEU and ROUGE metrics quantify the overlap between sequences of *n* words (or n-grams) in the generated output and the reference annotations (see Supplementary Materials). The BLEU score quantifies the precision of n-grams between the model output and reference annotations while also penalizing very short outputs compared to references, and the ROUGE score similarly quantifies the recall of n-grams, by comparing annotated snippets to model-generated answers to penalize the model when annotations are not included in outputs. Furthermore, the exact match F1-score between model outputs and annotated phrases was quantified to evaluate the model's ability in generating lexically-identical outputs. Of note is that the exact match metric is overly-strict, for example, *ER+: 2020* and *ER positive: 2020* are considered separate answers for exact match scores. Additionally, the accuracy of the best-performing model was quantified for 11 entity extraction tasks, and 20 relation extraction tasks on a random subset of half of the notes — 10 notes from breast cancer and 10 from pancreatic cancer — through review by an independent oncologist. Model outputs were presented to the oncologist as tables of the requested information by first parsing the GPT-4 model output automatically to populate the table. For example, the table for radiology test included columns for test name, datetime of the test, site of the test, reason for the test, and the test results. The expert assessment divided each output into three broad categories: correct, partially correct, and incorrect, which were aggregated to compute model accuracy. Partially correct or incorrect outputs were further subcategorized based on types of errors (see Supplementary Materials).

## Results

### Benchmarking dataset creation

Across 40 breast and pancreatic cancer progress notes, 9028 entities, 9986 entity attributes, and 5312 relationships were annotated, demonstrating the high density of clinically-relevant information in the complex medical oncology narratives. Patient demographics are presented in **Table 1**, and a sample of the annotated documentation is presented in **Figure 1**. Manual annotation was time-consuming; it took the oncology fellows 88 hours to annotate 27 documents, the student 50 hours to annotate 17 documents (4 of which were also annotated by a fellow), and the independent reviewer 116 hours to review potential errors in annotating all 40 documents. The mean inter-annotator agreement of entities — computed as the mean f1-score of overlap between their entity spans[32] — was 0.81, indicating high agreement.

**Table 1: Patient demographic distribution for the annotated data cohort comprising of 40 patients, additionally stratified by disease group.**

| Demographic Property | Category | Breast Cancer | Pancreatic Cancer | All |
|---|---|---|---|---|
| Race/Ethnicity | Native Hawaiian or Other Pacific Islander | 2 | 0 | 2 |
| | Latinx | 3 | 3 | 6 |
| | Native American or Alaska Native | 1 | 0 | 1 |
| | Southwest Asian and North African | 1 | 0 | 1 |
| | Black or African American | 4 | 4 | 8 |
| | Asian | 4 | 4 | 8 |
| | Multi-Race/Ethnicity | 0 | 3 | 3 |
| | Other | 1 | 2 | 3 |
| | White | 4 | 4 | 8 |
| | Unknown/Declined | 0 | 0 | 0 |
| Gender | Male | 0 | 10 | 10 |
| | Female | 20 | 10 | 30 |
| Age Group | Under 30 | 0 | 0 | 0 |
| | 31-40 | 6 | 0 | 6 |
| | 41-50 | 3 | 1 | 4 |
| | 51-60 | 6 | 4 | 10 |
| | 61-70 | 2 | 8 | 10 |
| | 71-80 | 2 | 6 | 8 |
| | 81-89 | 1 | 1 | 2 |

| | 89+ | | 0 | 0 | 0 |
|---|---|---|---|---|---|

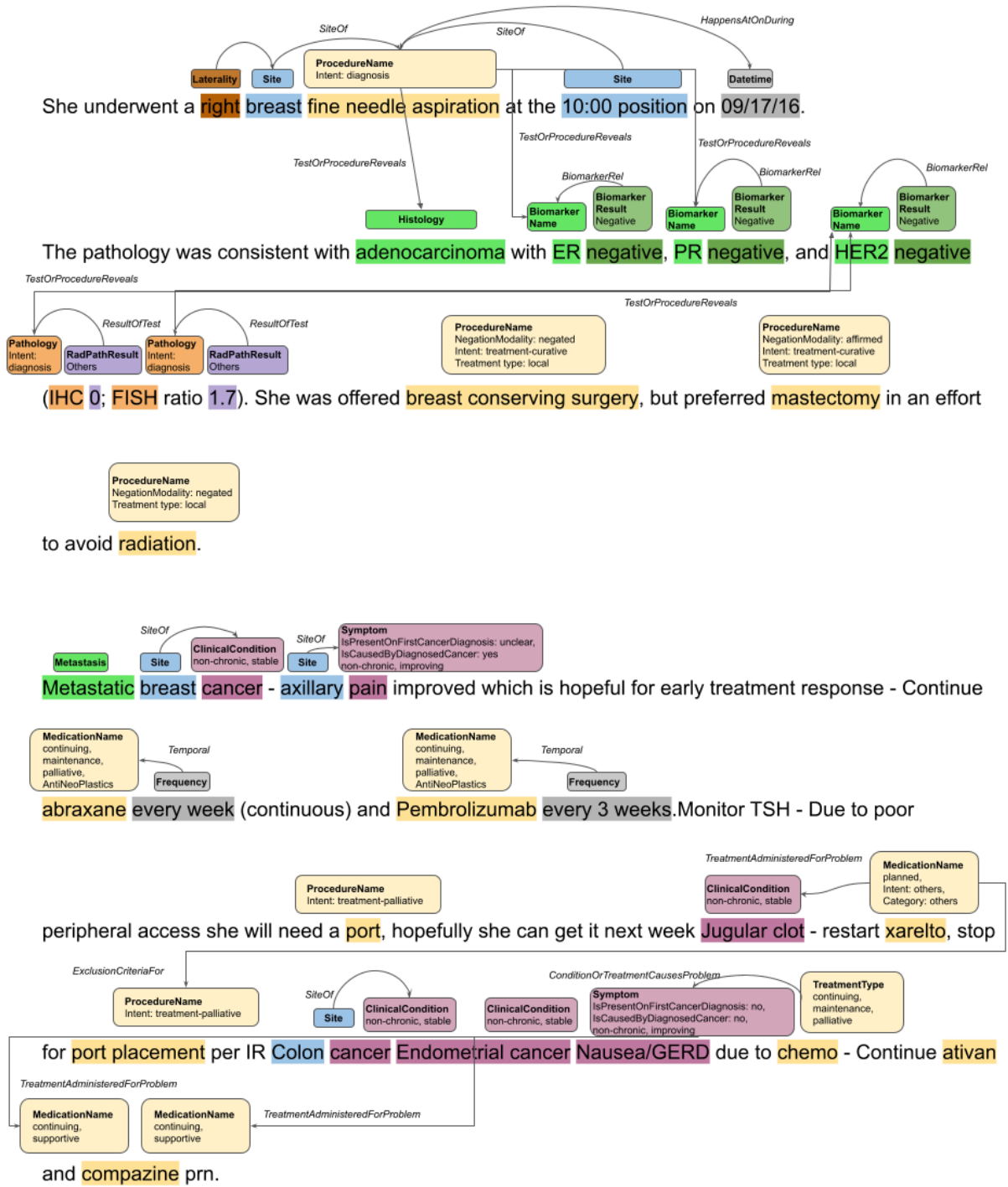

**Figure 1:** A sample of the annotated de-identified medical oncology progress note. The colored highlights refer to different types of entity spans within text. The arrows indicate the relations between the pair of entities linked. Within the box next to entity types, the corresponding modifier values for those entities are listed.

Description of initial cancer diagnosis, and disease and treatment progression were elaborately represented within the annotations (**Figure 2**). As anticipated, pancreatic cancer notes presented more palliative and supportive treatment entities, and more symptoms attributable to cancer. Conversely, breast cancer notes contained more diagnostic and staging tests, as well as laterality, lymph node involvement and biomarkers, pertaining to the more complex diagnostic and staging workup required for this disease. Temporal relations were common (1200 relations), as were indications of findings from a test or procedure (566 relations), and relations attributing adverse events to pre-existing conditions or treatments (232 relations).

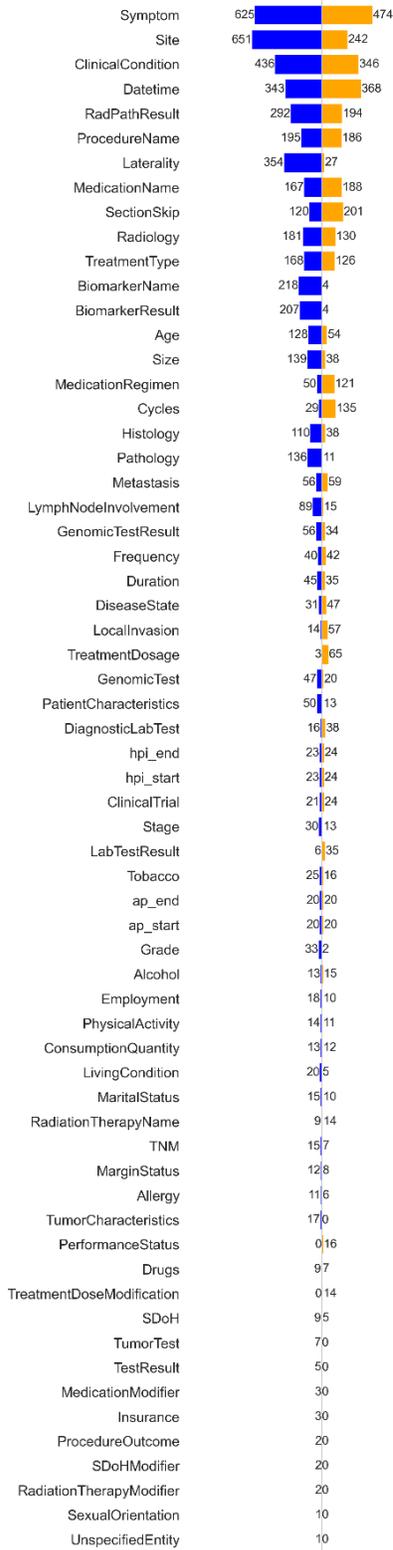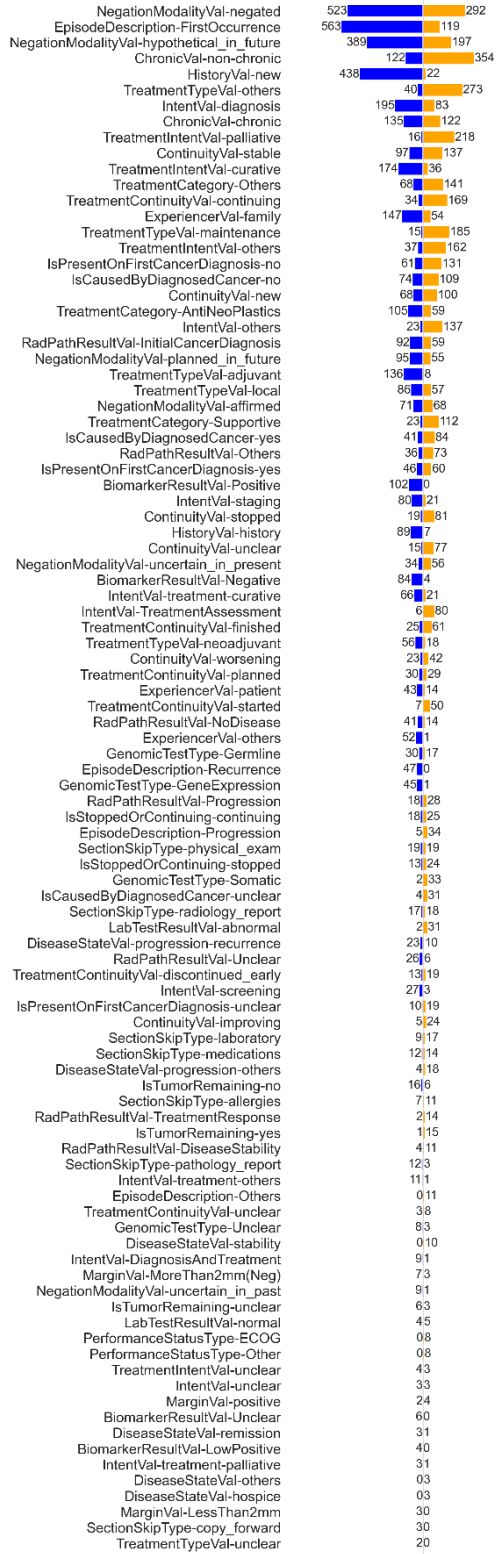

**(a)**

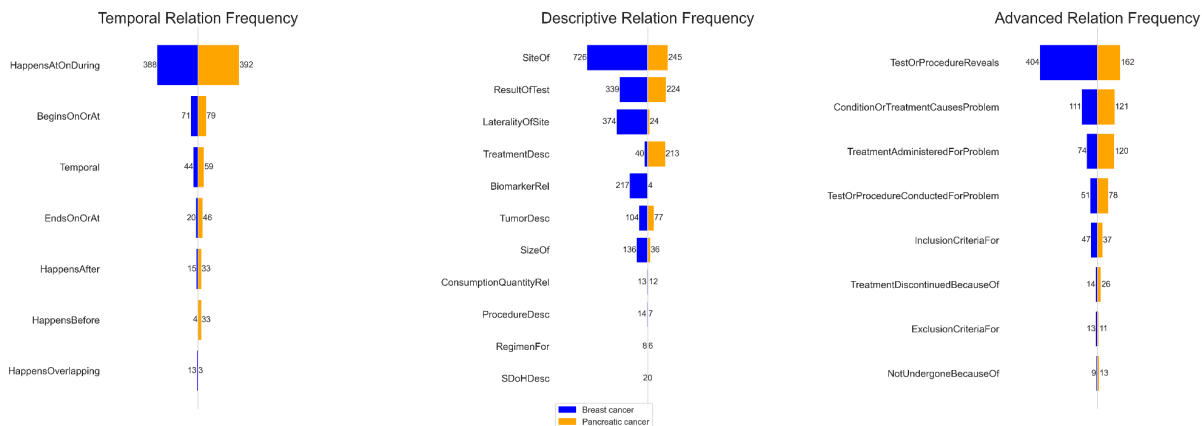

**(b)**

**Figure 2:** Distribution of: **(a)** the annotated entity mentions and the attribute values for these entities, and **(b)** relations in the annotated corpus of breast cancer medical oncology progress notes.

## Information Extraction

**GPT-4 outperforms the GPT-3.5 model and the FLAN-UL2 model**

The GPT-4 model performed better than GPT-3.5-turbo and FLAN-UL2 models (**Figure 3**), demonstrating an average BLEU score of 0.73, an average ROUGE score of 0.72, and an average exact-match F1 (EM-F1) score of 0.51 compared to the average BLEU, ROUGE, and EM-F1 scores of 0.61, 0.58, and 0.29 respectively for the GPT-3.5-turbo model, and 0.53, 0.27, and 0.06 respectively for the FLAN-UL2 model[3]. Performance differences were notable for tasks requiring advanced reasoning, for example, inferring symptoms present at the time of cancer diagnosis and hypothetical discussions of future medications.

---

[3] We additionally experimented with the clinical-T5-large model31, and LLaMA 7B, LLaMA 13B, and LLaMA-2 13B models32. However, we did not obtain reasonable outputs from these models for our task, presumably because they are not tuned on task descriptions, which is known to improve the instruction-following abilities of LLMs33.

**Best synthesis of tumor characteristics and medication history**

High scores — 0.95 BLEU, 0.93 ROUGE, 0.91 EM-F1 — were obtained by the GPT-4 model when extracting tumor grade paired with temporal information. High performances were also obtained in extracting cancer summary stage (0.85 BLEU, 0.80 ROUGE, 0.69 EM-F1), TNM stage (0.82 BLEU, 0.78 ROUGE, 0.71 EM-F1), and future medication (0.88 mean BLEU, 0.84 mean ROUGE, 0.70 mean EM-F1), suggesting promising capabilities in the automated extraction of these parameters. Furthermore, the GPT-4 model also demonstrated good performance with potential for further improvements in extracting genomics datetime and results (0.81 mean BLEU, 0.80 mean ROUGE, 0.68 mean EM-F1), radiology tests with their datetime (0.80 BLEU and ROUGE, 0.52 EM-F1), the prescribed medications with their start datetime, end datetime, current continuity status, potential adverse events, and adverse events experienced due to the medication (0.80 mean BLEU, 0.76 mean ROUGE, 0.57 mean EM-F1), symptoms with their datetime (0.71 BLEU, 0.74 ROUGE, 0.45 EM-F1), metastasis extraction (0.64 mean BLEU, 0.65 mean ROUGE, 0.44 EM-F1), and for identifying symptoms that occured due to cancer (0.67 BLEU, 0.75 ROUGE, 0.5 EM-F1). Lexical differences between annotated information and model outputs were common when extracting longer phrases, such as reasons for tests or test results, as compared to short responses like cancer grade or stage. The open-source FLAN-UL2 model and the GPT-3.5 model demonstrated higher lexical differences than the GPT-4 model, as evident from significantly lower EM F1-scores. When extracting histological subtypes and treatment-relevant tumor biomarkers, the models frequently provided more information than necessary, for example providing grade, stage, and biomarkers of a tumor in addition to the requested histological subtype. The poorest quantitative performance was obtained for procedure extraction (0.58 mean BLEU, 0.57 mean ROUGE, 0.33 EM-F1). GPT-4 model performance was comparable across the two cancer types, although the model demonstrated marginally better performance in medication and biomarker extraction for breast cancer, and in genomic and procedure extraction for pancreatic cancer (**Figure S2**, Supplementary materials). Furthermore, we did not find any significant differences in model performance across either male and female genders, or across different races and ethnicities for any of the three models for any of the three metrics, as computed with Kruskal-Wallis test with False discovery rate correction by Benjamini-Hochberg method (**Table S2, S3: Supplementary materials**).

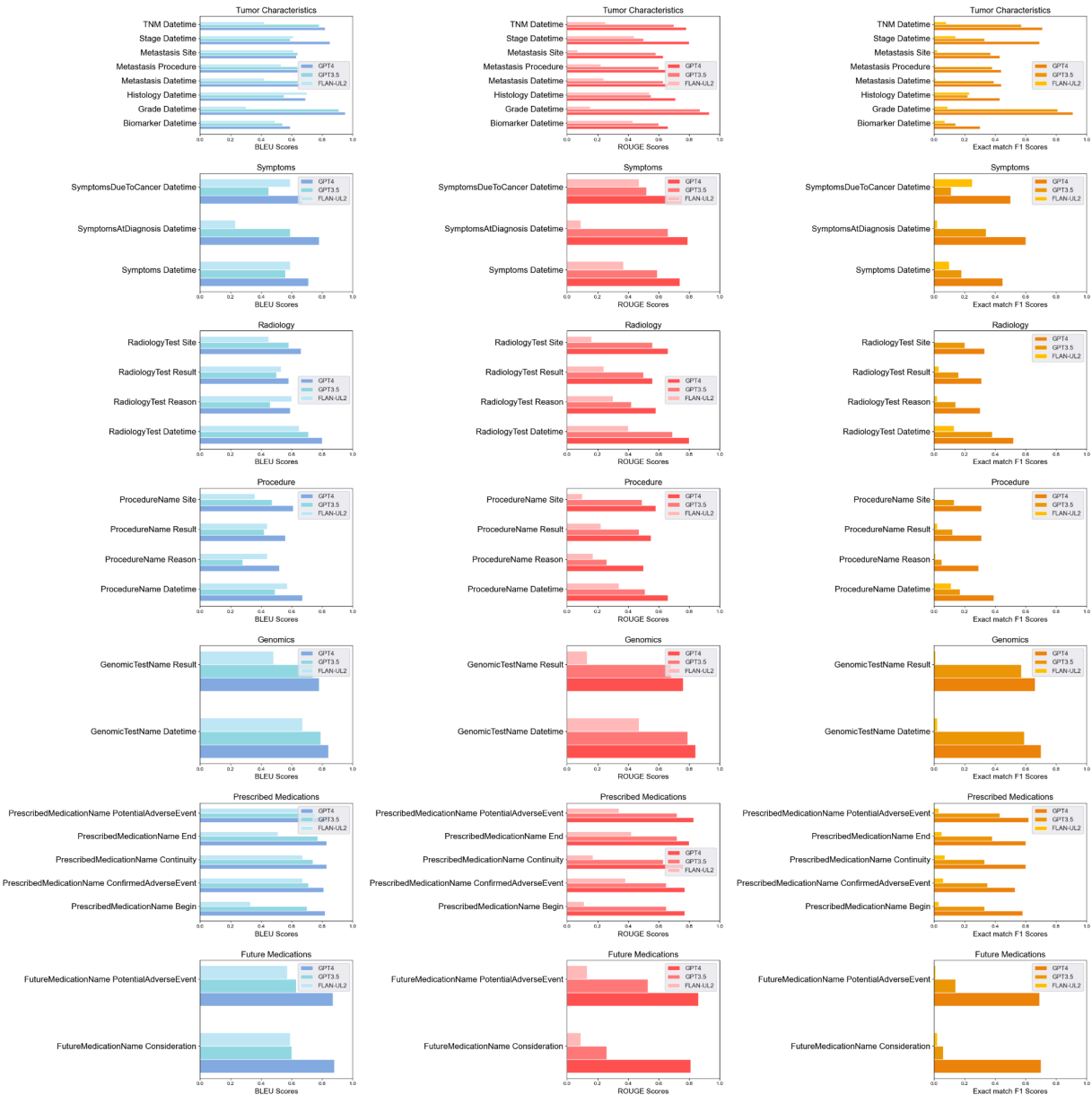

**Figure 3:** Mean BLEU scores (precision-focused; first column), ROUGE scores (recall-focused; second column), and exact-match F1-score (third column) for entity-relation extraction in different oncology tasks, grouped by the category of inference. *Entity1 Entity2* on the y-axis represents the performance for extracting the mention of entities *Entity1* and *Entity2* and thereby correctly inferring their relationship.

**Expert evaluation confirms superior oncologic information extraction ability**

A medical oncologist additionally evaluated GPT-4 model outputs on a subset of 10 breast cancer and 10 pancreatic cancer notes for the first cancer diagnosis date, symptoms, radiology tests, procedures, histology, metastasis, and future medications (**Table 2**). It took the oncologist nearly 90 hours to quantify GPT-4 model accuracy across 31 categories for the 20 notes. The expert evaluations showed that the GPT-4 model outputs were overall 68% accurate, with an additional 3% of the cases being correct, but missing some desired information and another 1% of the cases deemed to be uncertain due to linguistic ambiguity. These findings support the automated quantitative evaluations, highlighting the excellent oncologic information extraction ability of the GPT-4 model. The most common error (22%) were cases where the model produced output from note text, although it did not correspond to the requested information (*hallucinations 1; **Table 2: 3b***). Of the remaining errors, in 6% of the cases, the model returned *Unknown* instead of correct answers, and in 1% of the cases, the model fabricated information (*hallucinations 2; **Table 2: 3c***).

Partial correct answers with some missing information were most frequent for tumor histology, and in the results of radiology tests. Hallucinations from note text included incorrectly categorizing biomarkers, genomic tests, procedures, and radiology tests, and incorrect inferences of information like symptoms present at the time of cancer diagnosis or symptoms caused due to the diagnosed cancer. The model output information without direct references in text most frequently when inferring whether a medication was considered hypothetically or was planned for administration, and in identifying the reasons of tests and procedures, for example specifying that PET/CT was conducted to evaluate metabolic activity. Finally, the most common cases where the model output *Unknown* despite the correct information being present in the note included mentions of cancer histology, future medications, and symptoms due to cancer.

Table 2: Expert manual evaluation of GPT-4 outputs for a subset of inference categories on a subset of 10 breast cancer notes. Each note was first divided into the *History of Present Illness* and *Assessment and Plan* sections, thereby resulting in inference over 20 note snippets. Each cell represents the model outputs as a fraction of GPT-4 outputs in that category. When only one entity is mentioned, for example in *Procedure*, the scores represent the extraction of that entity by itself. When two entities are separated with a hyphen, for example in *Procedure - Datetime*, the scores represent an evaluation of correctly linking the entities together, for example pairing the *procedure* with the right *date or time of the procedure*. Partially correct answer categories are defined as follows — **(2a)**: The output contains more information than necessary, and the extra information is correct, **(2b)**: The output is correct, but some information is missing from the output, **(2c)**: The output contains more information than necessary, but the extra information is incorrect. Incorrect answer categories are defined as follows: **(3a)**: Independent expert reviewer determines that the note text is ambiguous, where either manual annotation or model answer could be considered correct, **(3b)**: Hallucinations 1- the model answers from the information mentioned in the note, but the answer is incorrect for the question asked., **(3c)**: Hallucinations 2 - the model fabricates information not discussed in the text, **(3d)**: Correct output is present in the input text, but the model returns unknown.

| Entity/ Relation | Inference Category | Sample size | Correct 1 | Partially correct 2a | 2b | 2c | Incorrect 3a | 3b | 3c | 3d |
|---|---|---|---|---|---|---|---|---|---|---|
| Entity | FirstCancerDiagnosis | 40 | 0.88 | 0.00 | 0.05 | 0.00 | 0.00 | 0.08 | 0.00 | 0.00 |
| | Symptoms | 140 | 0.84 | 0.04 | 0.01 | 0.00 | 0.00 | 0.06 | 0.00 | 0.04 |
| | SymptomsDueToCancer | 129 | 0.83 | 0.00 | 0.01 | 0.01 | 0.02 | 0.08 | 0.00 | 0.05 |
| | FutureMedication | 91 | 0.77 | 0.04 | 0.02 | 0.00 | 0.00 | 0.05 | 0.00 | 0.11 |
| | SymptomsAtDiagnosis | 63 | 0.76 | 0.00 | 0.00 | 0.00 | 0.00 | 0.24 | 0.00 | 0.00 |
| | Metastasis | 51 | 0.63 | 0.14 | 0.00 | 0.00 | 0.00 | 0.20 | 0.00 | 0.04 |
| | RadiologyTest | 116 | 0.53 | 0.27 | 0.00 | 0.00 | 0.00 | 0.13 | 0.00 | 0.07 |
| | Procedure | 107 | 0.51 | 0.15 | 0.01 | 0.01 | 0.01 | 0.27 | 0.01 | 0.03 |
| | Biomarker | 154 | 0.51 | 0.05 | 0.01 | 0.01 | 0.00 | 0.40 | 0.00 | 0.03 |
| | GenomicTestName | 84 | 0.50 | 0.00 | 0.00 | 0.00 | 0.00 | 0.46 | 0.00 | 0.04 |
| | Histology | 73 | 0.25 | 0.01 | 0.58 | 0.00 | 0.00 | 0.03 | 0.00 | 0.14 |
| Relation | Metastasis - Datetime | 51 | 0.90 | 0.00 | 0.00 | 0.00 | 0.00 | 0.06 | 0.00 | 0.04 |

| | | | | | | | | | |
|---|---|---|---|---|---|---|---|---|---|
| | SymptomsAtDiagnosis - Datetime | 63 | 0.87 | 0.00 | 0.00 | 0.00 | 0.00 | 0.11 | 0.00 0.02 |
| | Metastasis - Site | 51 | 0.86 | 0.02 | 0.00 | 0.00 | 0.00 | 0.08 | 0.00 0.04 |
| | Histology - Datetime | 72 | 0.86 | 0.00 | 0.00 | 0.00 | 0.00 | 0.00 | 0.00 0.14 |
| | Metastasis - Procedure | 51 | 0.80 | 0.06 | 0.04 | 0.00 | 0.00 | 0.06 | 0.00 0.04 |
| | Symptoms Datetime | 140 | 0.77 | 0.00 | 0.06 | 0.00 | 0.01 | 0.09 | 0.00 0.06 |
| | RadiologyTest - Datetime | 116 | 0.76 | 0.00 | 0.00 | 0.00 | 0.03 | 0.14 | 0.00 0.07 |
| | FutureMedication - PotentialAdverseEvent | 91 | 0.74 | 0.00 | 0.02 | 0.00 | 0.04 | 0.09 | 0.00 0.11 |
| | SymptomsDueToCancer - Datetime | 129 | 0.73 | 0.01 | 0.07 | 0.00 | 0.02 | 0.09 | 0.00 0.09 |
| | FutureMedication - Consideration | 91 | 0.73 | 0.00 | 0.00 | 0.00 | 0.00 | 0.09 | 0.08 0.11 |
| | Procedure - Datetime | 107 | 0.67 | 0.00 | 0.07 | 0.00 | 0.02 | 0.19 | 0.00 0.05 |
| | Biomarker - Datetime | 141 | 0.65 | 0.00 | 0.00 | 0.00 | 0.00 | 0.32 | 0.00 0.03 |
| | Procedure - Site | 107 | 0.61 | 0.00 | 0.01 | 0.00 | 0.01 | 0.34 | 0.01 0.03 |
| | RadiologyTest - Reason | 116 | 0.52 | 0.02 | 0.06 | 0.00 | 0.03 | 0.29 | 0.01 0.08 |
| | RadiologyTest - Result | 116 | 0.52 | 0.02 | 0.23 | 0.00 | 0.00 | 0.16 | 0.00 0.07 |
| | GenomicTestName - Datetime | 84 | 0.49 | 0.00 | 0.00 | 0.00 | 0.00 | 0.46 | 0.00 0.05 |
| | Procedure - Result | 107 | 0.49 | 0.00 | 0.15 | 0.00 | 0.00 | 0.30 | 0.04 0.03 |
| | GenomicTestName - Result | 84 | 0.46 | 0.00 | 0.00 | 0.00 | 0.00 | 0.46 | 0.04 0.04 |
| | Procedure - Reason | 107 | 0.42 | 0.04 | 0.07 | 0.00 | 0.03 | 0.42 | 0.00 0.03 |
| | RadiologyTest - Site | 116 | 0.40 | 0.01 | 0.01 | 0.00 | 0.03 | 0.47 | 0.01 0.09 |
| | **TOTAL/MEAN** | **2872** | **0.66** | 0.03 | 0.05 | 0.00 | 0.01 | 0.19 | 0.01 0.05 |

# Discussion

The schema presented in this manuscript, coupled with its associated annotated dataset, offers a robust benchmark for assessing LLM performance against human specialist curators in extracting complex details from medical oncology notes. Across 40 patient consultation notes, we manually annotated 9028 relevant entities, 9986 attributes, and 5312 relations, highlighting the information-dense nature of these notes. Our schema facilitated capturing of the nuanced rhetoric in medical oncology narratives, spanning information like family history, disease-relevant objective and temporal data, social determinants of health factors, causality between diagnoses, treatments, and symptoms, and treatment intent and response including potential and current adverse events. This new, rich dataset of real-world oncology progress notes will enable several follow-up advances in language models for oncology.

The dataset facilitated the benchmarking of the zero-shot capability of LLMs in oncologic history summarization. It demonstrated surprising zero-shot capability of the GPT-4 model in synthesizing oncologic history from the HPI and A&P sections, including tasks requiring advanced linguistic reasoning, such as extracting adverse events for prescribed medications and the reason for their prescription. The model, however, also showed room for improvements in causal inference, such as inferring whether a symptom was caused due to cancer. An open-source counterpart, the FLAN-UL2 model, demonstrated high precision but low recall, suggesting that it may be a promising alternative to the proprietary GPT-4 model if finetuned further on in-domain data.

Although current zero-shot performances are impressive because no task-specific fine-tuning was performed, the obtained accuracy may not be directly usable in clinical settings. Meanwhile, manual information extraction from notes is time-consuming, which contributes to an under-utilization of NLP in Electronic Health Record (EHR)-based observational research[33]. To obtain research-usable capability for oncology information extraction with minimal manual involvement, it is promising to explore strategies like few-shot learning, where few annotated examples are provided to the model to learn better[34], advanced prompt designs like chain-of-thought prompting[35] to benefit reasoning and selection-inference prompting[36] to first select the relevant entities before inferring more advanced relations and entity attributes, and in-domain fine-tuning for adapting open-source models to clinical domains. This dataset would be a critical resource for follow-up benchmarking studies, reducing the need for prior extensive domain-specific annotations for validating the performance of novel clinical LLMs and prompting strategies. Although the annotations in the dataset may be less verbose than model outputs, and annotator fatigue may contribute to additional errors, expert manual evaluations corroborated the findings of automated evaluations, demonstrating the reliability of the findings. Furthermore, although our study did not uncover any statistically significant disparities in model performance across gender and race/ethnicity, sample sizes in this study may be insufficient for this analysis, and there may be systematic biases in model performance that need to be studied further on larger datasets. Finally, small changes in prompts can result in a big impact on model performance. Further studies are needed to establish detailed guidelines with regards to prompt design and to quantify the impact of prompt engineering on model performance.

The current capability of LLMs in extracting tumor characteristics, medication, and adverse drug events demonstrated promise for enhanced post-approval real-world drug and device safety monitoring from unstructured data, automatically populating population-wide cancer registries, text-based cohort selection for EHR-based research studies, and speeding up the matching of patients to clinical trial criteria. Easier access to text data, potentially facilitated by LLMs in a human-in-the-loop setting, will improve research on patient outcomes and public health by providing evidence for better data-driven guidelines, and incorporating text-based variables in previously unutilized ways.

Although this dataset represents a small number of patients, the number of annotated sentences and oncologic concepts is large, making it comparable in sample size, but much larger in breadth, than existing benchmarking clinical NLP datasets. Furthermore, an additional set of 200 GPT-4-labeled notes would facilitate larger benchmarking studies through further expert analysis, and would also enable fine-tuning of open-source models with weak labels (model distillation). Finally, although we used the data from only two cancers within a single academic institution, the information representation and annotation schema was designed to be both cancer- and institution-agnostic, which will facilitate an extension of the analysis to large multi-center, multi-cancer studies to obtain generalizable conclusions.

# Conclusions

We successfully created a benchmarking dataset of forty expert-annotated breast and pancreatic cancer medical oncology notes by validating a new detailed schema for representing in-depth textual oncology-specific information, which is shared openly for further research. This dataset served as a test-bed to benchmark the zero-shot extraction capability of three LLMs: GPT-3.5-turbo, GPT-4, and FLAN-UL2. We found that the GPT-4 model holds the best performance with an average of 0.73 BLEU, 0.72 ROUGE, and 0.51 EM-F1 scores, highlighting the promising capability of language models to summarize oncologic patient history and plan without substantial supervision and aid future research and practice by reducing manual efforts. With further prompt engineering and model fine-tuning, combined with minimal manual supervision, LLMs will potentially be usable to extract important facts from cancer progress notes needed for clinical research, complex population management, and documenting quality patient care.

# Acknowledgements

This research would not have been possible with immense support from several people. The authors thank the UCSF AI Tiger Team, Academic Research Services, Research Information Technology, and the Chancellor's Task Force for Generative AI for their software development, analytical and technical support related to the use of Versa API gateway (the UCSF secure implementation of large language models and generative AI via API gateway), Versa chat (the chat user interface), and related data asset and services. We thank Boris Oskotsky, and the Wynton high-performance computing platform team for supporting high performance computing platforms that enable the use of large language models with de-identified patient data. We further thank Jennifer Creasman, Alysa Gonzales, Dalia Martinez, and Lakshmi Radhakrishnan for help with correcting clinical trial and gene name redactions. We thank Prof. Kirk Roberts for helpful discussions regarding frame semantics-based annotations of cancer notes, Prof.

Artuur Leeuwenberg for discussions about temporal relation annotation, and all members of the Butte lab for useful discussions in the internal presentations. Partial funding for this work is through the FDA grant U01FD005978 to the UCSF–Stanford Center of Excellence in Regulatory Sciences and Innovation (CERSI), through the NIH UL1 TR001872 grant to UCSF CTSI, through the National Cancer Institute of the National Institutes of Health under Award Number P30CA082103, and from a philanthropic gift from Priscilla Chan and Mark Zuckerberg. The content is solely the responsibility of the authors and does not necessarily represent the official views of the National Institutes of Health.

# Financial Disclosures and Conflicts of Interest

MS, VEK, DM, and TZ report no financial associations or conflicts of interest. BYM is a paid consultant for drug development at SandboxAQ with no conflicts of interest for this research. AJB is a co-founder and consultant to Personalis and NuMedii; consultant to Mango Tree Corporation, and in the recent past, Samsung, 10x Genomics, Helix, Pathway Genomics, and Verinata (Illumina); has served on paid advisory panels or boards for Geisinger Health, Regenstrief Institute, Gerson Lehman Group, AlphaSights, Covance, Novartis, Genentech, and Merck, and Roche; is a shareholder in Personalis and NuMedii; is a minor shareholder in Apple, Meta (Facebook), Alphabet (Google), Microsoft, Amazon, Snap, 10x Genomics, Illumina, Regeneron, Sanofi, Pfizer, Royalty Pharma, Moderna, Sutro, Doximity, BioNtech, Invitae, Pacific Biosciences, Editas Medicine, Nuna Health, Assay Depot, and Vet24seven, and several other non-health related companies and mutual funds; and has received honoraria and travel reimbursement for invited talks from Johnson and Johnson, Roche, Genentech, Pfizer, Merck, Lilly, Takeda, Varian, Mars, Siemens, Optum, Abbott, Celgene, AstraZeneca, AbbVie, Westat, and many academic institutions, medical or disease specific foundations and associations, and health systems. AJB receives royalty payments through Stanford University, for several patents and other disclosures licensed to NuMedii and Personalis. AJB's research has been funded by NIH, Peraton (as the prime on an NIH

# References

1. Savova GK, Danciu I, Alamudun F, et al. Use of Natural Language Processing to Extract Clinical Cancer Phenotypes from Electronic Medical Records. Cancer Research 2019;79(21):5463–70.
2. Nori H, King N, McKinney SM, Carignan D, Horvitz E. Capabilities of GPT-4 on Medical Challenge Problems.
3. Kung TH, Cheatham M, Medenilla A, et al. Performance of ChatGPT on USMLE: Potential for AI-assisted medical education using large language models. PLOS Digital Health 2023;2(2):e0000198.
4. Singhal K, Azizi S, Tu T, et al. Large language models encode clinical knowledge. Nature 2023;620(7972):172–80.
5. Lee P, Bubeck S, Petro J. Benefits, Limits, and Risks of GPT-4 as an AI Chatbot for Medicine. New England Journal of Medicine 2023;388(13):1233–9.
6. Leveraging GPT-4 for Post Hoc Transformation of Free-Text Radiology Reports into Structured Reporting: A Multilingual Feasibility Study [Internet]. [cited 2023 Apr 5];Available from: https://pubs.rsna.org/doi/epdf/10.1148/radiol.230725
7. Agrawal M, Hegselmann S, Lang H, Kim Y, Sontag D. Large language models are few-shot clinical information extractors [Internet]. In: Proceedings of the 2022 Conference on Empirical Methods in Natural Language Processing. Abu Dhabi, United Arab Emirates: Association for Computational Linguistics; 2022 [cited 2023 Mar 21]. p. 1998–2022.Available from: https://aclanthology.org/2022.emnlp-main.130
8. Haver HL, Ambinder EB, Bahl M, Oluyemi ET, Jeudy J, Yi PH. Appropriateness of Breast Cancer Prevention and Screening Recommendations Provided by ChatGPT. Radiology 2023;230424.
9. Alawad M, Yoon H-J, Tourassi GD. Coarse-to-fine multi-task training of convolutional neural networks for automated information extraction from cancer pathology reports. In: 2018 IEEE EMBS International Conference on Biomedical & Health Informatics (BHI). 2018. p. 218–21.
10. Breitenstein MK, Liu H, Maxwell KN, Pathak J, Zhang R. Electronic Health Record Phenotypes for Precision Medicine: Perspectives and Caveats From Treatment of Breast Cancer at a Single Institution. Clinical and Translational Science 2018;11(1):85–92.
11. Yala A, Barzilay R, Salama L, et al. Using machine learning to parse breast pathology reports. Breast Cancer Res Treat 2017;161(2):203–11.
12. Odisho * Anobel, Park B, Altieri N, et al. Pd58-09   extracting structured information from

pathology reports using natural language processing and machine learning. Journal of Urology 2019;201(Supplement 4):e1031–2.
13. Li Y, Luo Y-H, Wampfler JA, et al. Efficient and Accurate Extracting of Unstructured EHRs on Cancer Therapy Responses for the Development of RECIST Natural Language Processing Tools: Part I, the Corpus. JCO Clinical Cancer Informatics 2020;(4):383–91.
14. Altieri N, Park B, Olson M, DeNero J, Odisho AY, Yu B. Supervised line attention for tumor attribute classification from pathology reports: Higher performance with less data. Journal of Biomedical Informatics 2021;122:103872.
15. Zhou S, Wang N, Wang L, Liu H, Zhang R. CancerBERT: a cancer domain-specific language model for extracting breast cancer phenotypes from electronic health records. Journal of the American Medical Informatics Association 2022;29(7):1208–16.
16. Belenkaya R, Gurley MJ, Golozar A, et al. Extending the OMOP Common Data Model and Standardized Vocabularies to Support Observational Cancer Research. JCO Clinical Cancer Informatics 2021;(5):12–20.
17. Roberts K, Si Y, Gandhi A, Bernstam E. A FrameNet for Cancer Information in Clinical Narratives: Schema and Annotation [Internet]. In: Proceedings of the Eleventh International Conference on Language Resources and Evaluation (LREC 2018). Miyazaki, Japan: European Language Resources Association (ELRA); 2018. Available from: https://aclanthology.org/L18-1041
18. Datta S, Bernstam EV, Roberts K. A frame semantic overview of NLP-based information extraction for cancer-related EHR notes. Journal of Biomedical Informatics 2019;100:103301.
19. Mirbagheri E, Ahmadi M, Salmanian S. Common data elements of breast cancer for research databases: A systematic review. J Family Med Prim Care 2020;9(3):1296–301.
20. Datta S, Ulinski M, Godfrey-Stovall J, Khanpara S, Riascos-Castaneda RF, Roberts K. Rad-SpatialNet: A Frame-based Resource for Fine-Grained Spatial Relations in Radiology Reports [Internet]. In: Proceedings of the 12th Language Resources and Evaluation Conference. Marseille, France: European Language Resources Association; 2020. p. 2251–60.Available from: https://aclanthology.org/2020.lrec-1.274
21. Savova GK, Masanz JJ, Ogren PV, et al. Mayo clinical Text Analysis and Knowledge Extraction System (cTAKES): architecture, component evaluation and applications. J Am Med Inform Assoc 2010;17(5):507–13.
22. Savova GK, Tseytlin E, Finan SP, et al. DeepPhe - A Natural Language Processing System for Extracting Cancer Phenotypes from Clinical Records. Cancer Res 2017;77(21):e115–8.
23. Qi P, Zhang Y, Zhang Y, Bolton J, Manning CD. Stanza: A Python Natural Language Processing Toolkit for Many Human Languages [Internet]. 2020 [cited 2023 Dec 22];Available from: http://arxiv.org/abs/2003.07082
24. Zhang Y, Zhang Y, Qi P, Manning CD, Langlotz CP. Biomedical and clinical English model packages for the Stanza Python NLP library. Journal of the American Medical Informatics Association 2021;28(9):1892–9.
25. Radhakrishnan L, Schenk G, Muenzen K, et al. A certified de-identification system for all clinical text documents for information extraction at scale. JAMIA Open 2023;6(3):ooad045.
26. OpenAI. GPT-4 Technical Report [Internet]. 2023 [cited 2023 Apr 10];Available from: http://arxiv.org/abs/2303.08774
27. ChatGPT [Internet]. [cited 2023 Apr 10];Available from: https://chat.openai.com
28. A New Open Source Flan 20B with UL2 [Internet]. Yi Tay. [cited 2023 Apr 10];Available from: https://www.yitay.net/blog/flan-ul2-20b
29. News [Internet]. [cited 2023 Jul 28];Available from: https://physionet.org/news/post/415
30. Papineni K, Roukos S, Ward T, Zhu W-J. Bleu: a Method for Automatic Evaluation of Machine Translation [Internet]. In: Proceedings of the 40th Annual Meeting of the Association for Computational Linguistics. Philadelphia, Pennsylvania, USA: Association for Computational Linguistics; 2002 [cited 2023 Mar 20]. p. 311–8.Available from: https://aclanthology.org/P02-1040

31. Lin C-Y. ROUGE: A Package for Automatic Evaluation of Summaries [Internet]. In: Text Summarization Branches Out. Barcelona, Spain: Association for Computational Linguistics; 2004 [cited 2023 Mar 20]. p. 74–81.Available from: https://aclanthology.org/W04-1013
32. Hripcsak G, Rothschild AS. Agreement, the F-Measure, and Reliability in Information Retrieval. J Am Med Inform Assoc 2005;12(3):296–8.
33. Fu S, Wang L, Moon S, et al. Recommended practices and ethical considerations for natural language processing-assisted observational research: A scoping review. Clin Transl Sci 2023;16(3):398–411.
34. Brown T, Mann B, Ryder N, et al. Language Models are Few-Shot Learners [Internet]. In: Advances in Neural Information Processing Systems. Curran Associates, Inc.; 2020 [cited 2023 Apr 11]. p. 1877–901.Available from: https://proceedings.neurips.cc/paper/2020/hash/1457c0d6bfcb4967418bfb8ac142f64a-Abstract.html
35. Wei J, Wang X, Schuurmans D, et al. Chain-of-Thought Prompting Elicits Reasoning in Large Language Models [Internet]. 2022 [cited 2023 Jan 27]. Available from: https://openreview.net/forum?id=_VjQlMeSB_J
36. Creswell A, Shanahan M, Higgins I. Selection-Inference: Exploiting Large Language Models for Interpretable Logical Reasoning [Internet]. 2023 [cited 2023 Apr 10]. Available from: https://openreview.net/forum?id=3Pf3Wg6o-A4

# Supplementary Materials

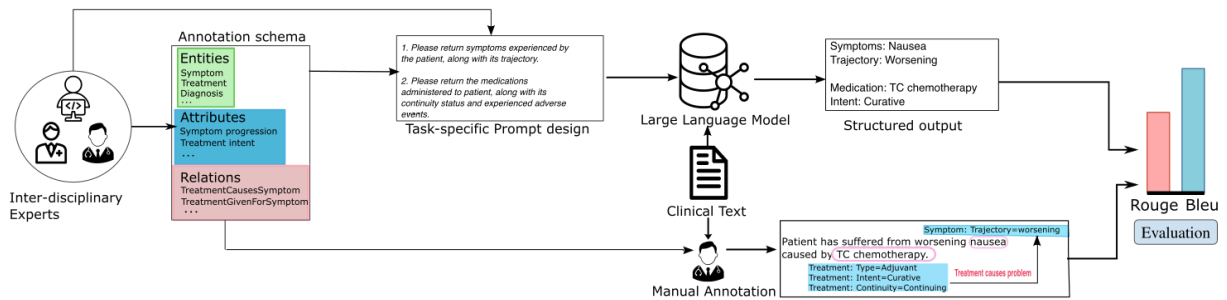

**Figure S1:** Workflow for developing an oncology information representation schema, annotating clinical text according to the schema, and evaluating large language models on specific tasks derived from the schema.

Annotations supported by the annotation schema

**Table S1:** List of annotation modalities, annotation buckets and annotation names supported by the newly developed schema to represent textual oncology information. Annotations are added within text through three annotation modalities: a) entities or phrases of specific type, b) attributes or modifiers of entities, and c) relations between entity pairs. These relations could either be (i) descriptive, for example relating a biomarker name to its results, (ii) temporal, for example indicating when was a test conducted, or (iii) advanced, for example, relating a treatment to adverse events caused due to it.

| **Annotation modality** | **Annotation bucket** | **Annotation name** |
|---|---|---|
| Entity | Temporal | Datetime |
| | | Frequency |
| | | Duration |
| | | Age |
| | PatientCharacteristics | Symptom |
| | | ClinicalCondition |
| | | Allergy |
| | | PerformanceStatus |
| | | SDoH |
| | | SDoH: Alcohol |
| | | SDoH: Drugs |
| | | SDoH: Tobacco |
| | | SDoH: PhysicalActivity |
| | | SDoH: Employment |
| | | SDoH: LivingCondition |
| | | SDoH: Insurance |

|  |  | SDoH: SexualOrientation |
|  |  | SDoH: MaritalStatus |
|  |  | SDoHModifier |
|  |  | SDoHModifier: ConsumptionQuantity |
|  | Location | Site |
|  |  | Laterality |
|  | TumorTest | GenomicTest |
|  |  | Pathology |
|  |  | Radiology |
|  |  | DiagnosticLabTest |
|  | TestResult | RadPathResult |
|  |  | GenomicTestResult |
|  |  | LabTestResult |
|  | TumorCharacteristics | Histology |
|  |  | Metastasis |
|  |  | LymphNodeInvolvement |
|  |  | Stage |
|  |  | TNM |
|  |  | Grade |
|  |  | Size |
|  |  | LocalInvasion |
|  |  | BiomarkerName |
|  |  | BiomarkerResult |
|  | Procedure | ProcedureName |
|  |  | ProcedureModifier |
|  |  | ProcedureModifier: ProcedureOutcome |
|  |  | ProcedureModifier: MarginStatus |

|  |  |  | MedicationName |
|---|---|---|---|
|  |  | Treatment | MedicationRegimen |
|  |  |  | MedicationModifier |
|  |  |  | MedicationModifier: Cycles |
|  |  |  | RadiationTherapyName |
|  |  |  | RadiationTherapyModifier |
|  |  |  | TreatmentDosage |
|  |  |  | TreatmentDoseModification |
|  |  |  | TreatmentType |
|  |  | ClinicalTrial | ClinicalTrial |
|  |  | DiseaseState | DiseaseState |
|  |  | SectionSkip | SectionSkip |
|  |  | UnspecifiedEntity | UnspecifiedEntity |
| Relation |  | Temporal | HappensAtOnDuring |
|  |  |  | BeginsOnOrAt |
|  |  |  | EndsOnOrAt |
|  |  |  | HappensBefore |
|  |  |  | HappensAfter |
|  |  |  | HappensOverlapping |
|  |  | Descriptive | ConsumptionQuantityRel |
|  |  |  | SDoHDesc |
|  |  |  | LateralityOfSite |
|  |  |  | SiteOf |
|  |  |  | SizeOf |
|  |  |  | ResultOfTest |
|  |  |  | BiomarkerRel |
|  |  |  | ProcedureDesc |

|  |  |  | TumorDesc |
|---|---|---|---|
|  |  |  | TreatmentDesc |
|  |  |  | RegimenFor |
|  |  | Advanced | TestOrProcedureReveals |
|  |  |  | TestOrProcedureConductedForProblem |
|  |  |  | TreatmentAdministeredForProblem |
|  |  |  | TreatmentDiscontinuedBecauseOf |
|  |  |  | ConditionOrTreatmentCausesProblem |
|  |  |  | NotUndergoneBecauseOf |
|  |  |  | InclusionCriteriaFor |
|  |  |  | ExclusionCriteriaFor |
| Attribute |  | NegationModalityVal | negated |
|  |  |  | affirmed |
|  |  |  | uncertain_in_present |
|  |  |  | uncertain_in_past |
|  |  |  | planned_in_future |
|  |  |  | hypothetical_in_future |
|  |  | PerformanceStatusType | ECOG |
|  |  |  | Karnofsky |
|  |  |  | Other |
|  |  | IsStoppedOrContinuing: SDoH | stopped |
|  |  |  | continuing |
|  |  | IsPresentOnFirstCancerDiagnosis: Symptom | yes |
|  |  |  | no |
|  |  |  | unclear |
|  |  | IsCausedByDiagnosedCancer: Symptom | yes |
|  |  |  | no |

| | | |
|---|---|---|
| | | unclear |
| | ChronicVal | chronic |
| | | non-chronic |
| | ContinuityVal | new |
| | | stable |
| | | improving |
| | | worsening |
| | | stopped |
| | | unclear |
| | ExperiencerVal | patient |
| | | family |
| | | others |
| | DiseaseStateVal | remission |
| | | progression-recurrence |
| | | progression-others |
| | | stability |
| | | others |
| | IntentVal | screening |
| | | staging |
| | | diagnosis |
| | | TreatmentAssessment |
| | | treatment-curative |
| | | treatment-palliative |
| | | treatment-others |
| | | DiagnosisAndTreatment |
| | | others |
| | | unclear |
| | RadPathResultVal | NoDisease |
| | | InitialCancerDiagnosis |
| | | Progression |
| | | DiseaseStability |
| | | TreatmentResponse |
| | | Mixed |
| | | Others |
| | | Unclear |
| | LabTestResultVal | normal |
| | | abnormal |
| | | unclear |
| | GenomicTestType | Germline |
| | | Somatic |
| | | GeneExpression |
| | | Others |
| | | Unclear |
| | MarginVal | positive |
| | | LessThan2mm |
| | | MoreThan2mm(Neg) |
| | | unclear |
| | HistoryVal | history |
| | | new |
| | EpisodeDescription | FirstOccurrence |
| | | Progression |
| | | Recurrence |
| | | Others |
|  | BiomarkerResultVal | Positive |
|  |  | Negative |
|  |  | LowPositive |
|  |  | Unclear |
|  | TreatmentContinuityVal | started |
|  |  | planned |
|  |  | continuing |
|  |  | finished |
|  |  | discontinued_early |
|  |  | unclear |
|  | CycleType | ongoing |
|  |  | completed |
|  |  | target |
|  | TreatmentTypeVal | neoadjuvant |
|  |  | adjuvant |
|  |  | maintenance |
|  |  | local |
|  |  | others |
|  |  | unclear |
|  | TreatmentIntentVal | curative |
|  |  | palliative |
|  |  | others |
|  |  | unclear |
|  | TreatmentCategory | AntiNeoPlastics |
|  |  | Supportive |
|  |  | Others |
|  | IsTumorRemaining | yes |

|  |  | no |
|  |  | unclear |
|  | SectionSkipType | physical_exam |
|  |  | medications |
|  |  | laboratory |
|  |  | allergies |
|  |  | radiology_report |
|  |  | pathology_report |
|  |  | copy_forward |
|  |  | others |

# Annotation guidelines

All the documents will be pre-highlighted with highlights 'Annotate' or 'Skip' to indicate whether a section should be annotated or can be skipped. The sections of interest may have been mentioned under different headers, which could have been missed by our rule-based pipeline for section detection. The best guesses should be made about the content of a paragraph or section based on their individual headings in the text.

## Sections to annotate

We will annotate the following sections in clinical notes:

- Sections describing the patient's history
    - History of present illness
    - Past Medical History
    - Past Surgical History
    - Subjective
    - Interval history
    - Oncology history
- Assessment and plan (Can also be mentioned as impression and plan)
- Review of systems: If this is mentioned as semi-structured data, ignore everything that is mentioned as N/A or not on file. Annotate the information that is known.
- Social history narrative: If this is mentioned as semi-structured data, ignore everything that is mentioned as N/A or not on file. Annotate the information that is known.
- Family history: If this is mentioned as semi-structured data, ignore everything that is mentioned

as N/A or not on file. Annotate the information that is known.

## Sections to skip

- The section 'Physical Examination" that describes the results of a physical exam (e.g. PHYSICAL EXAM, OBJECTIVE ASSESSMENT, etc.).
- The section containing semi-structured information about medications (e.g. CURRENT (OUTPATIENT) MEDICATIONS).
- The section containing semi-structured content about lab tests (e.g. Lab data, LABORATORY RESULTS, etc.).
- The section containing a review of allergies (e.g. ALLERGIES, Allergen Reactions, etc.).
- Radiology and Pathology sections (e.g. sections IMAGING, RADIOGRAPHIC AND PATHOLOGY RESULTS, SURGICAL PATHOLOGY REPORT, etc.)
    - Any radiology reports pasted within the note directly should be skipped as well.
- Copy forwarded information that has already been annotated.

The entire sections that are skipped should be annotated with the **SectionSkip** entity. The corresponding attribute *'SectionSkipType'* should be used to indicate the type of section that was skipped.

## Annotation schema details

The entity and relation annotations should be completed in the sequential order of documents. All the documents will be pre-highlighted with section headers 'Annotate' or 'Skip' entities of 3 types: problems, treatments, and tests, as recognized by publicly-available models. These are present as a guide rather than as the entities that we need to annotate ourselves.

The entities comprise everything of interest in the categories mentioned below in this schema. For all the entities, the minimum span that conveys the requisite information should be annotated. This would mean that adjectives that express additional information about an entity or an event should be retained, for example, "serious", however, articles such as "a", "an", "the" or other extraneous information should be excluded. To the extent feasible, a single span should be used for a single entity. Abbreviations should be annotated in an equivalent manner to their expanded forms.

All entities should specify details of negation and modality markers (*NegationModalityVal*), as well as temporality (with several temporal relations). The attribute *NegationModalityVal* takes a few different values: negated, affirmed, uncertain_in_present, uncertain_in_past, planned_in_future, and hypothetical_in_future. 'Affirmed' would be used by default whenever no other value has been set, so please take special care to indicate other values when relevant.

- Temporal: Several temporal relations can be used to relate any entities with these temporal marker entities: *HappensAtOnDuring*, *HappensBefore*, *HappensAfter*, *BeginsOnOrAt*, *EndsOnOrAt*, *HappensOverlapping*, *Temporal*.
    - **Datetime:** Includes date and time expressions, such as "December 2019", "12th

December", "2pm" as well as indirect references to time like "last year" or "2 months ago". A single span should be added for the entire DateTime expression. The minimum span that conveys the information should be used. For example, "last year" should be annotated instead of "in the last year".
- **Frequency:** Any details of frequency information, for example weekly, should be marked with this entity. Similar guidelines exist as for datetime entities for additional details.
- **Duration:** Used to mark temporal entities of type duration, for example, *10 hours*. Similar guidelines exist as for datetime entities for additional details.
- **Age:** Used to mark mentions of age. Similar guidelines exist as for datetime entities for additional details.

- **PatientCharacteristics** This generic category of PatientCharacteristics should be used for all mentions of relevant information other than Symptoms, ClinicalCondition, social determinants for health factors, and any other information explicitly listed as entities ahead. Some examples would be information about menopause, number of children, etc. that may be relevant for the cancer in question, or that may be relevant as an inclusion or exclusion criteria for a treatment or a clinical trial. *SiteOf* relation can be added for all symptoms and clinical conditions to add information about their site.
    - **Symptom:** This includes all mentions of symptoms and complaints that a patient presents with, for example, *fatigue, nausea, breathing difficulties*. Minimal entity spans that convey the information should be annotated, for example, *nausea* should be annotated instead of *the patient presents with nausea*. If the severity of the symptom is also mentioned, if relevant, should be included with the symptom entity (for example *minor breathing difficulties*).
        - The attribute *ContinuityVal* should be set to indicate whether it is a new symptom, a stable symptom, or whether the symptom has improved, worsened, or stopped.
        - The attribute *ChronicVal* should be set to indicate whether it is chronic or non-chronic.
        - Furthermore, the attribute *IsPresentOnFirstCancerDiagnosis* should be selected to indicate that the symptom was present when the patient was diagnosed with a new tumor for the first time. For this attribute, the value 'unclear' would be assumed by default if either yes or no is not indicated.
        - The attribute *IsCausedByDiagnosedCancer* should be set to indicate that the symptom was caused due to a cancer diagnosis. For this attribute, the value 'unclear' would be assumed by default if either yes or no is not indicated.
        - The attribute *NegationModalityVal* should be used to indicate any negation and modality markers jointly with all the entities.

        The relation *TreatmentAdministeredForProblem* should later be added between symptom entities and treatment names if the treatment was administered for a given symptom.

- The relation *TreatmentDiscontinueBecauseOf* should later be added to relate treatment entities with symptoms if the symptoms were the reason for discontinuing a treatment.

  ○ **ClinicalCondition:** This includes the diagnosis attached to the symptoms that the patient has presented with. This is usually stated by the provider and not the patient. Conditions like breast cancer and colon cancer are also included in this category. The attribute *ExperiencerVal* should be set to indicate whether the experiencer is the patient, family of the patient, or others. We can use 'patient' as experiencer by default, so please take special care to indicate this attribute if the experiencer is anyone other than the patient. Furthermore, the attribute *ChronicVal* should be set to indicate whether the clinical condition is a chronic condition or a non-chronic condition. The attribute *ContinuityVal* should additionally be set to indicate whether it is a new condition, continuing condition, whether the condition has progressed, or it has stopped. As always, *NegationModalityVal* should be used to highlight negation/modality.

    The relation *TreatmentAdministeredForProblem* should later be added between clinical condition entities and treatment names if the treatment was administered for a given clinical condition.

    The relation *TreatmentDiscontinueBecauseOf* should also be added to relate treatment entities with a clinical condition if it was a reason for discontinuing a treatment.

    If a clinical condition causes a symptom or another condition, the relation *ConditionOrTreatmentCausesProblem* should be added.

  ○ **Allergy:** Any existence of allergy within the sections that are annotated should be marked as the Allergy entity. The same relations as Symptom and ClinicalCondition are also present for allergy. The attribute *ContinuityVal* is used to determine whether it is new, stable, improving, worsening, or stopped. As always, *NegationModalityVal* should be used to highlight negation/modality.

  ○ **PerformanceStatus:** Mentions of performance status, such as the ECOG score should be marked with this entity. The attribute *PerformanceStatusType* should be additionally set to provide the type of the status.

  ○ **SDoH:** This includes different social determinants for health factors. The broad-level category of SDoH should be used when a more specific category (Alcohol, Drugs, Tobacco, ConsumptionQuantity, PhysicalActivity, Employment, LivingCondition, Insurance, SexualOrientation, and MaritalStatus) is not present. The attributes *NegationModalityVal* and *ExperiencerVal* should be used as needed. The attributes *IsStoppedOrContinuing* (default: continuing) should be used to indicate whether these values are in the past (for example, the patient has stopped drinking), or continuing in the present (the patient drinks).

- - - **Alcohol:** Mentions of alcohol use.
    - **Drugs:** Mentions of drug use.
    - **Tobacco:** Mentions of tobacco consumption in any form. Smokeless tobacco would be included in this category in addition to smoking. Vaping, if it contains nicotine, would be included.
    - **PhysicalActivity:** All phrases describing physical activity should be marked as such, for example, swimming, walking 2 miles per day, etc.
    - **Employment:** Mentions of employment status. This can include mentions of the profession, for example, "doctor", or mentions of employment status, for example "employed in the private sector". The entity spans should cover the details of the profession when available, but no extra information.
    - **LivingCondition:** Information about the living condition of the patient, for example, "lives alone", "lives with family", "homeless", etc.
    - **Insurance:** Any information about the insurance, for example, information about the insurance plan, or the presence or absence of insurance.
      The relation *TreatmentDiscontinueBecauseOf* should later be added to relate treatment entities with insurance status if it was a reason for discontinuing the treatment.
    - **SexualOrientation:** Any mention of the patient's sexual orientation.
    - **MaritalStatus:** Any mentions of the patient's marital status.
    - **SDoH Modifier:** Any important modifiers for SDoH, particularly those not listed explicitly, should be annotated as this entity. For example: type of tobacco use. These entities should further be related with the corresponding SDoH with an *SDoHDesc* relation.
      - **ConsumptionQuantity:** Any mentions of consumption quantity for alcohol, drugs, or tobacco, for example, 3 packs per day or 1 pint of beer a day. Any mentions of consumption frequency should also be encoded within the same entity.
      - A relation between this entity and the corresponding Alcohol/Drugs/Tobacco entity should be added at the subsequent step (*ConsumptionQuantityRel*).
- Location: The attribute *NegationModalityVal* should be used for the following entities as needed.
  - **Site**: This includes the site (organ/body part) of a test, the site of the tumor on the body, or the site of a surgical procedure, for example 'breast'. If laterality is present, it should be annotated separately. For example, in "left breast", only "breast" should be annotated as the site. When the exact site is mentioned, for example, *10 o'clock from the nipple*, the entire minimal span that conveys the information should be annotated. If the exact location of a specimen on an organ is not present, only the broader topography should be annotated, in line with the ICD-O3 topography guidelines [here](#) (or in the following guide:
  https://apps.who.int/iris/bitstream/handle/10665/96612/9789241548496_eng.pdf;

starting on page 44; breast cancer on page 54).

Site is later related to the name of the test/procedure/tumor that it is a site of using the relation *SiteOf*.

If the site is mentioned in an ambiguous manner, such as 'left lumpectomy', please annotate 'left' as the site and not laterality.

- **Laterality:** Laterality of the site where either a test was conducted, where the tumor was found, or where a surgical procedure was performed. for example 'left' in 'left breast'.

  Laterality should later be related to the corresponding site (*LateralityOfSite*).

- **TumorTest:** This includes any tests conducted to either screen or confirm whether a tumor is present. The generic category of TumorTest will be used only when a test specification is not precisely either covered under Radiology, Pathology, GenomicTest, or DiagnosticLabTest. The site/laterality of the test should be annotated as separate entities and should not be a part of this same entity called TumorTest.

  The attribute value under *IntentVal* should be used for each TumorTest to indicate whether the test was conducted for screening, staging, treatment assessment, or other reasons. Similarly, the attribute value under *VenueVal* should be used to indicate whether the test happened at UCSF, at other locations, or is unclear (default). If the tests would be conducted in the future, the attribute value 'hypothetical_in_future' should be used for *NegationModalityVal*. If the tests were negated, for example, "did not undergo a biopsy", "biopsy" should be annotated, and the negation attribute should be selected under *NegationModalityVal*. The attribute *ExperiencerVal* should be used to indicate whether the experiencer is the patient (default), family, or others.

  (These relations are also discussed later at the end of the document: ) The site of the test should be added with the relation *SiteOf*.

  If a test was conducted in response to an existing problem, tumor characteristics, or disease progression, the relation *TestOrProcedureConductedForProblem* should be used to relate them.

  The details of the test results should be related to this entity using the relation *ResultOfTest* described later.

  If a test reveals a malignant tumor, clinical condition, or disease progression, the test name should be related to the relevant entity using the relation *TestOrProcedureReveals*.

  - **Pathology:** Any mentions of pathology tests conducted should be annotated. If the type of biopsy is specified, for example, "core needle biopsy", it should be annotated as a part of the same entity.
    - Specific cases:

- FISH (Fluorescence in situ hybridization): Please label this as a Pathology test
    - **Radiology:** The annotations should follow similar strategies as that of the Pathology annotations. All screening, as well as diagnostic tests, should be annotated. For example, "Mammography", "Ultrasound", "US", 'MRI', 'CT' should all be annotated under this category.
    - **DiagnosticLabTest:** This includes lab tests that are specifically conducted to diagnose a tumor. Some examples include PSA, CEA, CA99, AFP, etc.
    - **GenomicTest:** This entity should be added for all types of gene tests. Additionally, the attribute *GenomicTestType* should be added to indicate whether the genomic test is of types germline test, somatic test, gene expression test (for example Oncotype Dx or Mammaprint), others, or the type unclear from text. Remember that for any hypothetical discussions of tests to be conducted in the future, the *NegationModalityVal* attribute value "hypothetical_in_future" should additionally be used.
- **TestResult**: This is the generic category for the tumor test result. This category should be reserved for any tests except pathology, radiology, genomic test, or diagnostic lab test. The attribute *ExperiencerVal* should be used to indicate whether the test result is about the patient (default), family, or someone else, and the attribute *NegationModalityVal* should be used to indicate the negation/modality status.

  The relation *ResultOfTest* should be used to relate the test name with its corresponding result (this result can either be the generic category of TestResult, or a more specific category of RadPathResult, GenomicTestResult, or LabTestResult).

    - **RadPathResult:** This category is for results related to both pathology and radiology tests. Attribute value *RadPathResultVal* should be set in addition to highlighting the corresponding entity to indicate whether the test result refers to the no disease, initial disease diagnosis, stable disease, disease progression, treatment response, mixed response (for example one tumor is improving, the other is worsening), others, or is unclear.
    - **GenomicTestResult:** This entity is for all results of genomic tests. The attributes *NegationModalityVal*, *ExperiencerVal,* and *GenomicTestType* should be set to indicate the corresponding relevant values.
    - **LabTestResult**: This entity is reserved for annotating the results of DiagnosticLabTest. In addition, the attribute value *LabTestResultVal* should be selected to indicate whether the test is 'normal', 'abnormal', or 'unclear'.
- **TumorCharacteristics**: This includes any modifiers not covered in the list mentioned below. For all the tumor characteristics in this list, the following attributes should be set: the attribute *ExperiencerVal* should be set to indicate whether the patient (default), family members, or others have these tumor characteristics. Modifiers of old tumors should be accompanied with the value "history" of the attribute *HistoryVal* (default value: new). Similarly, all TumorModifier entities should also indicate an *EpisodeDescription* attribute after deciphering whether it is the histology at the "FirstOccurrence" (default), "progression", "recurrence" or "others". Any negation and modality should be expressed with the *NegationModalityVal* attribute (default: affirmed).

The relation *TestOrProcedureReveals* would later be added for any tumor properties that have been identified with the help of a test or a procedure.

The relation *TestOrProcedureConductedForProblem* should later be added for any test or procedure that was conducted for any tumor characteristics.

The relation *SiteOf* should be used to relate site entities to the corresponding tumor properties.

The relation *TreatmentAdministeredForProblem* should later be added to describe any relations between treatments administered for any of the tumor characteristics.

- **Histology**: This includes the histology of the tumor either revealed by a test or the histology of a tumor in a patient's history. Unless a more specific description is available, the tumor histology should also follow the ICD-O3 histology, which can be referenced [here](#) (and by using 'ICD-O3' 'Search' option after selecting the relevant text span in BRAT).
- **Metastasis:** Entity to annotate the mentions of tumor metastasis (to other locations than lymph nodes) in text. If metastasis is negated, the corresponding attribute for *NegationModalityVal* should be selected. Relations should later be added to describe details related to the metastasis event, including the Datetime of metastasis (*Temporal*), site of metastasis (*SiteOf*), and the corresponding tumor histology, if any (*MetastasisDesc*).
- **LymphNodeInvolvement**: Entity to highlight mentions of lymph node involvement in text. The number of lymph nodes involved should be included as a part of this entity if mentioned, same for micrometastasis or any other relevant specific detail. The site of lymph nodes should not be included as a part of this entity, but should separately be annotated as Site, and later related to this entity using the *'SiteOf'* relation. Note that we will not annotate the lymph nodes examined at this stage, only the lymph nodes involved.
- **Stage:** Numeric mentions of the tumor stage, as well as mentions like "early" stage, should be annotated here, for example, 'IV' in "Stage IV". The prefix "Stage" should not be annotated, and only the number or the stage description should be. If the stage is mentioned as a range, for example 'Stage I-IIA', 'I-IIA' should be annotated.
- **TNM:** The TNM stages should be annotated here. This includes all TNM stages either as a single entity or as individual entities. For example, 'T2a', 'pT1N0M0', 'pN0', 'pMX', 'pT1N1M0(i+)', 'ypT1', 'cT1N1M1' should all be annotated. Note that we include both pathological and clinical stages, along with details of TNM such as prefixes for change of stage after therapy (y), as well as suffixes about the procedure, such as (i+).
- **Grade:** Tumor grade. This includes both numeric values such as '1' (do not include the

prefix 'grade'), as well as textual mentions such as 'low' in 'low grade', 'intermediate' in 'intermediate grade', etc.
- **Size:** The size of the tumor, along with the corresponding units of measurement, should be annotated. For example '7*9*5mm' should be annotated as a single entity.
    - Size can be for any pathological/surgical/radiological procedure and then related to that procedure through *TestOrProcedureReveals*, in addition to the descriptive relation *SizeOf*.
- **LocalInvasion**: Annotate all mentions of whether there is a local invasion. This would include LVI (microscopic), as well as other types of invasion, for example, 'invades chest walls' etc.
- **BiomarkerName:** Any names of cancer biomarkers should be annotated. For the sake of generalization, our definition of a biomarker is very loose. For example, we would include 'ER', 'PR' as well as 'HER-2' as biomarker names for breast cancer. PSA for prostate cancer would be annotated as a DiagnosticLabTest and not a biomarker. Please annotate only the final test results if the test was conducted multiple times. If the results are present as history, please highlight it as the corresponding attribute value *'HistoryVal'*. Please note that any terms indicating the results, such as '+' in 'ER+' should be annotated as a separate result and would not be a part of the BiomarkerName entity. Any details of the test should not be included in the annotation span and instead should be labeled as pathology test.

    Immunohistochemistry stains should be labeled here under biomarker. The name of the gene or protein that is being tested for should be the selected entity

    These results would be related with the corresponding entities using the *BiomarkerRel* temporalrelation.

- **BiomarkerResult:** The results of a biomarker, for example positive, negative, equivocal, '+', '-' etc. should be annotated as the BiomarkerResult. Any percentages, if mentioned, should be included as a part of the result, for example, '+ 100%'. The attribute value *BiomarkerResultVal* should additionally be used to indicate a positive, negative, low positive, or unclear result.

    A relation *BiomarkerRel* between the BiomarkerName and the BiomarkerResult should subsequently be added to link different biomarker names with biomarker results.

- Procedure:
    - **ProcedureName:** Name of a procedure. This includes all types of diagnostic procedures, screening procedures, as well as treatment-related surgeries, such as 'lumpectomy', 'mastectomy', 'colonoscopy', etc. All procedures get the attribute *IntentVal* to indicate the intent of the procedure: whether it is for screening, staging, diagnosis, treatment assessment, treatment (curative, palliative, or others), diagnosing as well as treatment, or others. *TreatmentTypeVal* entity should be set to indicate whether it is an adjuvant

procedure, neoadjuvant, maintenance, local, or other procedures (if the procedure is related to treatments). If the procedure is being discussed for the future, the corresponding *NegationModalityVal* attribute should be set. *IsTumorRemaining* entity should be used to indicate whether any tumor was remaining after the procedure or not if the procedure was to remove the tumor (i.e. treatment procedure). This attribute can be left blank if the procedure was not a procedure done with the intent of treating the patient. The attribute *TreatmentCategory* should be used to indicate whether it is an antineoplastic treatment (default), supportive treatment, or others. The attribute *ExperiencerVal* should be set to indicate whether the patient (default), family members, or others have these tumor characteristics.

If a procedure was conducted in response to an existing problem, tumor characteristics, or because of disease progression, the relation *TestOrProcedureConductedForProblem* should be added.

If a procedure reveals a malignant tumor, clinical condition, or disease progression, the ProcedureName name should be related to the relevant entity using the relation *TestOrProcedureReveals*.

If any problem such as a symptom or a clinical condition was caused by a procedure, the relation *ConditionOrTreatmentCausesProblem* should be added.

- **ProcedureModifier:** Any modifiers for the procedure that are not described by either the outcome or the margin status should be annotated as this generic entity. The attributes *NegationModalityVal* and *ExperiencerVal* should be set as described earlier for all procedure modifiers.

  For all the modifiers (including ProcedureOutcome and MarginStatus), the relation *ProcedureDesc* should be added to describe which procedure are they modifiers for.

  - **ProcedureOutcome**: Any outcome or result of a procedure, such as results of a biopsy that is not pathology can be annotated as the ProcedureOutcome entity. This can further be related to the ProcedureName entity.
  - **MarginStatus:** Value of the surgical margins, along with corresponding units if any. If there are multiple resections, only the final margin should be annotated. The MarginStatus entity takes the attribute *MarginVal* to indicate whether the margin is positive, more than 2mm (negative), less than 2 mm, or unclear. Only the final margin should be annotated in the case of multiple mentions.
- Treatment: Treatments, broadly categorized under the following categories: MedicationName, MedicationRegimen, RadiationTherapyName, or TreatmentType. For all of these entities, a few attributes should be set. The attribute *TreatmentContinuityVal* should be used to indicate whether a treatment was started (immediately), planned (in the future; not immediately), finished, discontinued early, or is continuing currently. The default value "started" would be used if not indicated otherwise. Similarly, under the attribute *TreatmentIntentVal*, we should add the intent of therapy: whether it was curative, palliative, or others. The default value of "curative" would be used if not indicated otherwise. The attribute *TreatmentTypeVal* should be used to indicate whether the treatment is neoadjuvant, adjuvant, maintenance, local, or others. A relation should be added between all treatment entities and their corresponding modifiers (*TreatmentDesc*). The attributes *NegationModalityVal* and *ExperiencerVal* should be used as needed with all of the following entities. Relation (*TreatmentAdministeredForProblem*) should be added between treatment names and the clinical condition, symptoms, tumor characteristics, or disease progression entities that they have been administered for.

- **MedicationName**: This includes the name of all cancer-specific medication therapies for all modes of administration, including chemotherapy, hormone therapy as well as immunotherapy (but not radiation therapy). Every individual medication name should be annotated as a separate entity. Medications administered for managing secondary symptoms should be annotated as "supportive medications. There is a separate category for medication modifiers like dosage, so only the name should be included under this entity. An attribute *TreatmentCategory* should be set to indicate whether it is an antineoplastic medication, supportive medication, or others (others can be skipped unless it seems to be particularly important).

  If any problem such as a symptom or a clinical condition was caused because of a medication, the relation *ConditionOrTreatmentCausesProblem* should be added between the problem and the medication name.

  If a medication was discontinued because of a symptom, clinical condition, disease progression, insurance, or hospice, the relation *TreatmentDiscontinuedBecauseOf* should be added between the reason and the medication name.

- **MedicationRegimen**: If mentioned explicitly in the text, add the name of the regimen for a medication. If the text also mentions the name of medications under this regimen, a relation (*RegimenForName*) should be added between the MedicationRegimen and the MedicationName entities.

  If any problem such as a symptom or a clinical condition was caused because of a medication regimen, the relation *ConditionOrTreatmentCausesProblem* should be added between the problem and the medication regimen.

  If a medication regimen was discontinued because of a symptom, clinical condition, disease progression, insurance, or hospice, the relation *TreatmentDiscontinuedBecauseOf* should be added between the reason and the medication regimen.

- **RadiationTherapyName:** This includes the name of the radiation therapy.

If any problem such as a symptom or a clinical condition was caused because of radiation therapy, the relation *ConditionOrTreatmentCausesProblem* should be added between the problem and the radiation therapy name.

If radiation therapy was discontinued because of a symptom, clinical condition, disease progression, insurance, or hospice, the relation *TreatmentDiscontinuedBecauseOf* should be added between the reason and the radiation therapy name.

- **TreatmentDosage**: This includes dosage for medications as well as RadiationTherapy. The units for dosage should also be included. If the discussions are related to dose reduction instead of explicit mention of dosage, for example 20%, that should be annotated as the next entity, 'TreatmentDoseModification'. A relation should later be added between the TreatmentDosage and their corresponding name (*TreatmentDesc*).
- **TreatmentDoseModification:** As mentioned earlier, this includes modifications to the usual dosages, for example, '20%', '200%', or '3/4$^{th}$'. A relation should later be added between the TreatmentDoseModification and their corresponding name (*TreatmentDesc*).
- **TreatmentType:** The type of therapy, for example, adjuvant therapy, chemotherapy, etc. An attribute value to indicate whether they are neoadjuvant, adjuvant, maintenance, local, or another type of therapy is also provided as *'TreatmentTypeVal'*.
- **MedicationModifier:** Any modifier apart from those mentioned next, as well as apart from dosage, and dose modification, should be annotated as the generic entity MedicationModifier. A relation should later be added between all modifiers and their corresponding name (*TreatmentDesc*). The duration and frequency can be annotated as *Datetime* expressions, with the right type set under the *DatetimeVal* attribute.
    - **Cycles:** Number of cycles that a medication is administered for, for example, '5' in '5 cycles'. Please only highlight the number, and none of the terms that talk about the cycle itself. If it is a range, please annotate the entire range.
- **RadiationTherapyModifier:** Any modifiers for the RadiationTherapy apart from dosage, dose modification, duration, and type (adjuvant, neoadjuvant, maintenance, local, others), should be highlighted as this entity. A relation should later be added between the RadiationTherapyModifier and RadiationTherapyName (*TreatmentDesc*).

- **ClinicalTrial:** Names of clinical trials discussed in the text should be annotated. These names should be modified with the *NegationModalityVal* attribute as needed to indicate either any hypothetical discussions or to indicate whether the patient participated in the trial or not. Similarly, *ExperienceVal* should be used to indicate whether the discussed clinical trial is in reference to the patient or others.

    Furthermore, the relations *'InclusionCriteriaFor'* and *'ExclusionCriteriaFor'* should be used to relate all other entities to the trial if they qualify as inclusion and exclusion criteria respectively. There is an additional option to add temporal relations for ClinicalTrial to indicate the DateTime of the trial.

- **DiseaseState**: The entity should be marked for any mentions of disease state, such as remission, progression, hospice, etc. The relevant attribute value of *DiseaseStateVal* should be marked to indicate the type of the DiseaseState. If the concept is negated, then this entity should be marked, along with its corresponding *NegationModalityVal* attribute value. Any discussions about anyone other than the patient should be highlighted with the attribute *ExperienceVal*.

  Many relations are possible for this entity, such as whether a test or procedure was conducted due to a disease state (*TestOrProcedureConductedForProblem*), whether a test or procedure revealed disease state (*TestOrProcedureReveals*), whether a treatment was discontinued due to a disease state (*TreatmentDiscontinuedBecauseOf*), whether a treatment was administered because of a disease state *(TreatmentAdministeredForProblem),* or to indicate *InclusionCriteria, ExclusionCriteria, and Temporal relations.* These relations should be added in the second phase.

- **UnspecifiedEntity:** This category will be used whenever none of the existing entities match what should be annotated, but the information is relevant and should be added to the schema. Please use the comments box to describe this unspecific entity: i) what it is, and ii) why it is relevant.

**Relation annotations**

- **Temporal relation for all entities:** The following relations are allowed between the (Datetime) entity and all other entities in our data. Different types of temporal relations allowed are described next. The detailed description of these relations can be accessed in the [THYME annotation guidelines](#) under 'TLINK' descriptions (Section 6.2). These relations are symmetric and can be added in either direction.
    - Event **BeginsOnOrAt** Datetime (BEGINS)
    - Event **EndsOnOrAt** Datetime (ENDS)
    - Event **HappensAtOnDuring** Datetime (DURING/CONTAINS)
    - Event **HappensBefore** Datetime (BEFORE)
    - Event **HappensAfter** Datetime (AFTER)
    - Event **HappensOverlapping** Datetime (OVERLAP). This refers to the situation when an event happens overlapping a datetime mention. These are more common in relations between two events though.
- **Descriptive relations:**
    - **For SDoH:**
        - **ConsumptionQuantityRel:** Relation between the (ConsumptionQuantity) entity and (Alcohol, Drug or Tobacco) entities.
        - **SDoHDesc**: Relation between any SDoHModifier and the SDoH entities.
    - **Relations related to site and laterality:**
        - **LateralityOfSite:** Relation from the (Laterality) entity to the (Site) entity, which indicates the laterality for the mentioned site.
        - **SiteOf:** Relation from the (Site) entity to any of the test name or result, procedure name or result, tumor modifier, and problem entities (TumorTest,

Radiology, Pathology, GenomicTest, DiagnosticLabTest, ProcedureName, TestResult, RadPathResult, GenomicTestResult, LabTestResult, Histology, Metastasis, LymphNodeInvolvement, Stage, TNM, Grade, Size, LocalInvasion, BiomarkerName, TumorCharacteristics, Symptom, Clinical condition, Allergy, MedicationName, MedicationRegimen, RadiationTherapyName, TreatmentType, DiseaseState).
Note that if multiple sites of tests exist and different sites have different results, this relation should be linked to the test result in addition to the test name.
We have allowed this relation for all tumor modifiers, instead of only histology, so that they can be used if needed. For example, there can be cases where the test conducted isn't mentioned again when (site of) tumor modifiers are mentioned, or the site for the test isn't mentioned, although the site for a biomarker is present in a note.
- **SizeOf:** Relation between a (Size) entity and other entities indicating what is it the size of: (TreatmentDosage, TreatmentDoseModification, MedicationModifier, Cycles, AdministrationMode, RadiationTherapyModifier, TestResult, RadPathResult, Symptom, ClinicalCondition, Allergy, ProcedureName).
- **ResultOfTest:** Relation from results of tests, which includes either of these entities: (TestResult, RadPathResult, GenomicTestResult, LabTestResult) to the entities describing diagnostic tests conducted for finding tumors (TumorTest, Radiology, Pathology, GenomicTest, DiagnosticLabTest, ProcedureName). Note that relations with procedure name are added as this type only if the procedure provides a RadPathResult, a GenomicTestResult, or a LabTestResult. Any other procedure results would be annotated as the ProcedureOutcome entity, and related with the ProcedureDesc relation instead.
- **ProcedureDesc:** This relates a procedure that was conducted (either surgical or diagnostic) and its modifiers. The relation is added between entity pairs (ProcedureName) and the entities (MarginStatus, ProcedureOutcome, ProcedureModifier).
- **BiomarkerRel:** Between (BiomarkerName) and its result (BiomarkerResult).
- **TumorDesc:** A bidirectional relation between the entities (TumorCharacteristics, ClinicalCondition) (used to indicate terms like cancer) to all tumor modifiers (Histology|Metastasis|LymphNodeInvolvement|Stage|TNM|Grade|Size|LocalInvasion|BiomarkerName). This relation should be added particularly if a test or procedure was not conducted for this tumor and there exists a dangling relation between tumor modifiers and its parent otherwise.
- **TreatmentDesc:** Between Treatment name entities (MedicationName, MedicationRegimen, RadiationTherapyName, TreatmentType) and Treatment modifiers (TreatmentDosage, TreatmentDoseModification, MedicationModifier, Cycles, AdministrationMode, RadiationTherapyModifier).
- **RegimenFor:** Relation between (MedicationRegimen) and (MedicationName or TreatmentType) entities.
- **Advanced relations:**

- **TestOrProcedureConductedForProblem:** Relation that indicates that a test was conducted because of a symptom or a clinical condition that the patient presented with. This relation is added between the entities (TumorTest, Pathology, Radiology, GenomicTest, DiagnosticLabTest, ProcedureName) and (Symptom, ClinicalCondition, Allergy, DiseaseState, Histology, Metastasis, LymphNodeInvolvement, Stage, TNM, Grade, Size, LocalInvasion, BiomarkerName, TumorCharacteristics).
- **TestOrProcedureReveals:** Relation between different test entities (TumorTest, Pathology, Radiology, GenomicTest, DiagnosticLabTest, ProcedureName) and the entities that have revealed a clinical condition, disease progression, or tumor characteristics (Histology, Metastasis, LymphNodeInvolvement, Stage, TNM, Grade, Size, LocalInvasion, BiomarkerName, TumorCharacteristics, DiseaseState, Symptom, ClinicalCondition, Allergy).
- **TreatmentDiscontinuedBecauseOf:** Relation that indicates the reason for discontinuing treatment, if any. This relation is added from the entity group (MedicationName, MedicationRegimen, RadiationTherapyName, TreatmentDosage, TreatmentDoseModification, MedicationModifier, Cycles, AdministrationMode, RadiationTherapyModifier) to the reasons, which encompass most other entities.
- **ConditionOrTreatmentCausesProblem:** This relation is added to indicate that any treatment caused a new symptom or clinical condition. The relation would be added between entity pairs (MedicationName, MedicationRegimen, RadiationTherapyName, TreatmentType, ProcedureName, ClinicalCondition) and (ClinicalCondition, Symptom, Allergy).
- **TreatmentAdministeredForProblem:** This relation is added to indicate that the treatment (MedicationName, MedicationRegimen, RadiationTherapyName, TreatmentType) was administered for a given reason (symptom, clinical condition, allergy, or PatientCharacteristics, DiseaseState, Histology, Metastasis, LymphNodeInvolvement, Stage, TNM, Grade, Size, LocalInvasion, BiomarkerName, TumorCharacteristics, TestResult, RadPathResult, GenomicTestResult, LabTestResult, PerformanceStatus). If any of the modifiers is not mentioned along with treatment, then no need to relate it across several notes.
- **NotUndergoneBecauseOf:** Relation to indicate that a treatment, test, procedure, or trial was not undergone because of any reason such as insurance, distance (annotated as PatientCharacteristics), disease progression, etc. This relation can be added between the entities (MedicationName|MedicationRegimen|RadiationTherapyName|TreatmentType|TumorTest|Radiology|Pathology|GenomicTest|DiagnosticLabTest|ProcedureName|ClinicalTrial) and **MOST** other entities (PatientCharacteristics|Symptom|ClinicalCondition|Allergy|SDoH|Alcohol|Drugs|Tobacco|PhysicalActivity|Employment|LivingCondition|Insurance|SexualOrientation|MaritalStatus|TumorTest|Radiology|Pathology|GenomicTest|DiagnosticLabTest|TestResult|RadPathResult|GenomicTestResult|LabTestResult|Histology|Metastasis|LymphNodeInvolvement|Stage|TNM|Grade|Size|LocalInvasion|BiomarkerName|TumorCharacteristics|Cli

- nicalTrial|MedicationName|MedicationRegimen|RadiationTherapyName|TreatmentType|TreatmentDosage|TreatmentDoseModification|MedicationModifier|Cycles|AdministrationMode|RadiationTherapyModifier|ProcedureName|ProcedureModifier|MarginStatus|ProcedureOutcome|DiseaseState).
  - **Relations about clinical trials:**
    - **InclusionCriteriaFor**: This should be used to relate ALL other entities to either of the entities (ClinicalTrial, MedicationName, MedicationRegimen, RadiationTherapyName, TreatmentType, ProcedureName) if they qualify as the inclusion criteria for this trial/treatment/procedure. If multiple entities make up a single inclusion criterion, all of them should be related to the trial name individually. The only exception is when there exists a transitive relation between the entities, for example when medication dosage is linked to medication name, we can only link medication name to the trial.
    - **ExclusionCriteriaFor**: This should be used to relate ALL other entities to either of the entities (ClinicalTrial, MedicationName, MedicationRegimen, RadiationTherapyName, TreatmentType, ProcedureName) if they qualify as the inclusion criteria for this trial/treatment/procedure.

## Quantitative Evaluation Metrics

**BLEU metric**

The BLEU metric was designed for automated evaluation of text pairs to quantify the overlap between tokens, or words, in these pairs. The metric provides a modified precision score by computing the overlap between position-independent n-grams of a candidate string clipped to their maximum occurrence in the reference string (in our case the lower-cased generated model output, for example, *headache of mild nature, 11th december 2022*) and a set of correct reference strings (in our case the lower-cased manual annotations of the requested information, for example, {*mild headache: 11th december 2022, fever: 11th december 2022*}), as follows:

$$p_n = \frac{\sum_{C \in \{\text{Candidates}\}} \sum_{\text{n-gram} \in C} \text{Count}_{\text{clip}}(\text{n-gram})}{\sum_{C' \in \{\text{Candidates}\}} \sum_{\text{n-gram}' \in C'} \text{Count}(\text{n-gram}')}$$

The scores are averaged over the entire corpus to get an overall quality estimate. A brevity penalty is added to penalize candidate strings that are too short compared to the reference strings, since being a precision-focused metric, short answers would inevitably lead to a high BLEU score. The final scores are computed as such:

$$\log \text{BLEU} = \min(1 - \frac{\text{reference length}}{\text{candidate length}}, 0) + \sum_{n=1}^{N} w_n \log p_n$$

The length of n-grams, or the number of words in matched phrases, is a hyperparameter, which we set to be 4; $w_n$ is a weight, set to a uniform value 0.25. To avoid penalizing annotations that are shorter than 4 words, smoothing is added, which disregards higher order n-gram comparisons if the reference annotations are shorter. Of note is that 1.0 BLEU score indicates a perfect match, and that BLEU can only quantify lexical overlap, and does not measure semantic overlap between strings.

**ROUGE-n metric**

Similar to the BLEU metric, the ROUGE-n metric was also designed for automated evaluation of text pairs by quantifying an overlap between tokens of the candidate string and a set of correct reference strings. Instead of being precision-focused, this metric is recall-focused, and quantifies the proportion of reference token n-grams that are also covered by the output text, as follows:

$$\text{ROUGE-n} = \frac{\sum_{s \in \text{Reference Summaries}} \sum_{n-gram \in \text{S}} \text{Count}_{\text{match}}(n-gram)}{\sum_{s \in \text{Reference Summaries}} \sum_{n-gram \in \text{S}} \text{Count}(n-gram)}$$

We quantify overlap by setting *n* to be 1, which quantifies single word overlap between the references and the candidate.

**BLEU and ROUGE metric example**

BLEU and ROUGE scores of 1.0 reflect a perfect overlap between model output and reference annotations (high performance), and a score of 0 refers to no overlap (poor performance). For example, if the model output for requested radiology test and datetime is "*chest MRI: december 2015*", and the manual annotations refer to it as "*MRI: 12th december 2015*", the BLEU-4 score would be 0.45 and the ROUGE-1 score will be 0.75. BLEU score is low because the model provided the site *chest* in addition to the annotated information. ROUGE score is lower than 1.0 since the model did not include the date, *12th*, in its output although it has been annotated.

## Scheme for expert manual evaluation

The expert evaluation categorized GPT-4 output into 3 categories (correct, partially correct, incorrect). Partially correct and incorrect answers were further categorized into specific modes according to the following scheme:

1. The output is entirely correct
2. The output is partially correct
   a. The output contains more information than necessary, and the extra information is correct. For example, the model outputs the histology as *stage 2 IDC*, although the expected answer is only *IDC*, and *stage 2* is the right stage for this cancer.

b. The output is correct, but some information is missing from the output. For example, the model outputs only *MRI: right breast*; when the MRI was conducted for both right and left breasts.

c. The output contains more information than necessary, but the extra information is incorrect. For example, the model outputs the histology as *stage 2 IDC*, although the expected answer is only *IDC*, and *stage 2* is the incorrect stage for this cancer.

3. The output is incorrect

    a. Independent expert reviewer determines that the note text is ambiguous, where either manual annotation or model answer could be considered correct.

    b. Hallucinations 1- the model answers from the information mentioned in the note, but the answer is incorrect for the question asked. For example, an *MRI* that was conducted is reported as a *procedure* instead of a *radiology test*.

    c. Hallucinations 2 - the model fabricates information not discussed in the text. For example, even though no *MRI* was conducted, the model outputs that *MRI* was conducted for the patient.

    d. Correct output is present in the input text, but the model returns *unknown*.

## Additional Results

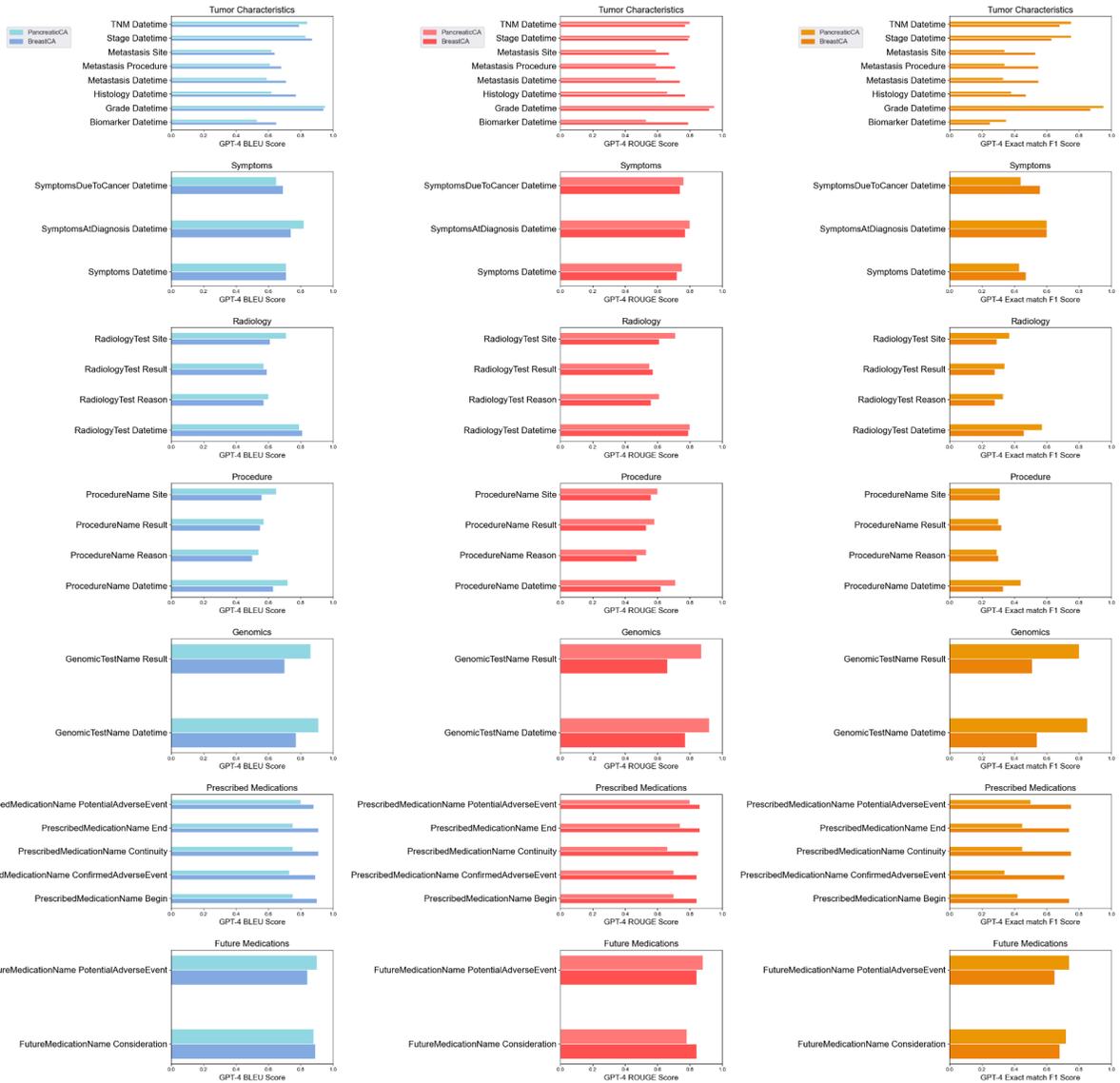

**Figure S2:** GPT-4 performance (BLEU scores, ROUGE scores, and exact match F1 scores) stratified by disease group.

**Table S2: Kruskal-Wallis test statistics for comparing model performance across the included genders variables for each information extraction task and metric. The q values are computed after correcting False Discovery Rate with Benjamini–Hochberg method.**

| Task | Model | Metric | Kruskal F-statistic | Kruskal p-value | Kruskal q-value |
|---|---|---|---|---|---|
| Symptom Datetime | gpt-4 | BLEU4 | 0.87 | 0.35 | 0.64 |
| Symptom Datetime | gpt-4 | ROUGE1 | 1.16 | 0.28 | 0.61 |
| Symptom Datetime | gpt-4 | EM_F1 | 2.12 | 0.14 | 0.44 |
| RadiologyTest Datetime | gpt-4 | BLEU4 | 2.73 | 0.10 | 0.36 |
| RadiologyTest Datetime | gpt-4 | ROUGE1 | 5.14 | 0.02 | 0.21 |
| RadiologyTest Datetime | gpt-4 | EM_F1 | 6.93 | 0.01 | 0.14 |
| RadiologyTest Site | gpt-4 | BLEU4 | 5.68 | 0.02 | 0.19 |
| RadiologyTest Site | gpt-4 | ROUGE1 | 9.47 | 0.00 | 0.08 |
| RadiologyTest Site | gpt-4 | EM_F1 | 3.88 | 0.05 | 0.33 |
| RadiologyTest Reason | gpt-4 | BLEU4 | 5.42 | 0.02 | 0.20 |
| RadiologyTest Reason | gpt-4 | ROUGE1 | 8.42 | 0.00 | 0.10 |
| RadiologyTest Reason | gpt-4 | EM_F1 | 3.87 | 0.05 | 0.33 |
| RadiologyTest Result | gpt-4 | BLEU4 | 4.68 | 0.03 | 0.24 |
| RadiologyTest Result | gpt-4 | ROUGE1 | 2.21 | 0.14 | 0.44 |
| RadiologyTest Result | gpt-4 | EM_F1 | 5.45 | 0.02 | 0.20 |
| ProcedureName Datetime | gpt-4 | BLEU4 | 1.20 | 0.27 | 0.60 |
| ProcedureName Datetime | gpt-4 | ROUGE1 | 2.86 | 0.09 | 0.34 |
| ProcedureName Datetime | gpt-4 | EM_F1 | 2.57 | 0.11 | 0.38 |
| ProcedureName Site | gpt-4 | BLEU4 | 2.41 | 0.12 | 0.40 |
| ProcedureName Site | gpt-4 | ROUGE1 | 1.48 | 0.22 | 0.57 |
| ProcedureName Site | gpt-4 | EM_F1 | 0.06 | 0.81 | 0.93 |
| ProcedureName Reason | gpt-4 | BLEU4 | 0.97 | 0.33 | 0.62 |
| ProcedureName Reason | gpt-4 | ROUGE1 | 3.15 | 0.08 | 0.34 |
| ProcedureName Reason | gpt-4 | EM_F1 | 1.02 | 0.31 | 0.61 |
| ProcedureName Result | gpt-4 | BLEU4 | 0.52 | 0.47 | 0.72 |
| ProcedureName Result | gpt-4 | ROUGE1 | 1.37 | 0.24 | 0.60 |
| ProcedureName Result | gpt-4 | EM_F1 | 1.13 | 0.29 | 0.61 |
| GenomicTestName Datetime | gpt-4 | BLEU4 | 1.21 | 0.27 | 0.60 |
| GenomicTestName Datetime | gpt-4 | ROUGE1 | 0.88 | 0.35 | 0.64 |

| Task | Model | Metric | Kruskal F-statistic | Kruskal p-value | Kruskal q-value |
|---|---|---|---|---|---|
| GenomicTestName Datetime | gpt-4 | EM_F1 | 1.79 | 0.18 | 0.51 |
| GenomicTestName Result | gpt-4 | BLEU4 | 0.78 | 0.38 | 0.66 |
| GenomicTestName Result | gpt-4 | ROUGE1 | 1.48 | 0.22 | 0.57 |
| GenomicTestName Result | gpt-4 | EM_F1 | 1.27 | 0.26 | 0.60 |
| Biomarker Datetime | gpt-4 | BLEU4 | 1.73 | 0.19 | 0.52 |
| Biomarker Datetime | gpt-4 | ROUGE1 | 2.69 | 0.10 | 0.36 |
| Biomarker Datetime | gpt-4 | EM_F1 | 0.02 | 0.88 | 0.94 |
| Histology Datetime | gpt-4 | BLEU4 | 3.11 | 0.08 | 0.34 |
| Histology Datetime | gpt-4 | ROUGE1 | 3.09 | 0.08 | 0.34 |
| Histology Datetime | gpt-4 | EM_F1 | 2.52 | 0.11 | 0.39 |
| Metastasis Site | gpt-4 | BLEU4 | 0.22 | 0.64 | 0.83 |
| Metastasis Site | gpt-4 | ROUGE1 | 0.00 | 0.99 | 1.00 |
| Metastasis Site | gpt-4 | EM_F1 | 0.73 | 0.39 | 0.67 |
| Metastasis Procedure | gpt-4 | BLEU4 | 0.19 | 0.66 | 0.83 |
| Metastasis Procedure | gpt-4 | ROUGE1 | 0.06 | 0.81 | 0.93 |
| Metastasis Procedure | gpt-4 | EM_F1 | 1.02 | 0.31 | 0.61 |
| Metastasis Datetime | gpt-4 | BLEU4 | 0.09 | 0.76 | 0.91 |
| Metastasis Datetime | gpt-4 | ROUGE1 | 0.23 | 0.63 | 0.83 |
| Metastasis Datetime | gpt-4 | EM_F1 | 0.87 | 0.35 | 0.64 |
| Stage Datetime | gpt-4 | BLEU4 | 0.06 | 0.81 | 0.93 |
| Stage Datetime | gpt-4 | ROUGE1 | 0.04 | 0.83 | 0.93 |
| Stage Datetime | gpt-4 | EM_F1 | 0.04 | 0.85 | 0.93 |
| TNM Datetime | gpt-4 | BLEU4 | 0.20 | 0.66 | 0.83 |
| TNM Datetime | gpt-4 | ROUGE1 | 0.10 | 0.75 | 0.91 |
| Grade Datetime | gpt-4 | BLEU4 | 2.13 | 0.14 | 0.44 |
| Grade Datetime | gpt-4 | ROUGE1 | 3.32 | 0.07 | 0.34 |
| Grade Datetime | gpt-4 | EM_F1 | 3.33 | 0.07 | 0.34 |
| MedicationName Begin | gpt-4 | BLEU4 | 1.92 | 0.17 | 0.48 |
| MedicationName Begin | gpt-4 | ROUGE1 | 1.09 | 0.30 | 0.61 |

| Task | Model | Metric | Kruskal F-statistic | Kruskal p-value | Kruskal q-value |
|---|---|---|---|---|---|
| MedicationName Begin | gpt-4 | EM_F1 | 1.98 | 0.16 | 0.47 |
| MedicationName End | gpt-4 | BLEU4 | 6.46 | 0.01 | 0.17 |
| MedicationName End | gpt-4 | ROUGE1 | 2.87 | 0.09 | 0.34 |
| MedicationName End | gpt-4 | EM_F1 | 3.39 | 0.07 | 0.34 |
| MedicationName Reason | gpt-4 | BLEU4 | 3.19 | 0.07 | 0.34 |
| MedicationName Reason | gpt-4 | ROUGE1 | 2.14 | 0.14 | 0.44 |
| MedicationName Reason | gpt-4 | EM_F1 | 3.06 | 0.08 | 0.34 |
| MedicationName Continuity | gpt-4 | BLEU4 | 5.33 | 0.02 | 0.20 |
| MedicationName Continuity | gpt-4 | ROUGE1 | 1.78 | 0.18 | 0.51 |
| MedicationName Continuity | gpt-4 | EM_F1 | 3.07 | 0.08 | 0.34 |
| MedicationName ConfirmedAdvEvent | gpt-4 | BLEU4 | 1.45 | 0.23 | 0.57 |
| MedicationName ConfirmedAdvEvent | gpt-4 | ROUGE1 | 0.03 | 0.86 | 0.93 |
| MedicationName ConfirmedAdvEvent | gpt-4 | EM_F1 | 0.53 | 0.47 | 0.72 |
| MedicationName PotentialAdvEvent | gpt-4 | BLEU4 | 1.08 | 0.30 | 0.61 |
| MedicationName PotentialAdvEvent | gpt-4 | ROUGE1 | 0.53 | 0.46 | 0.72 |
| MedicationName PotentialAdvEvent | gpt-4 | EM_F1 | 1.33 | 0.25 | 0.60 |
| MedicationName Consideration | gpt-4 | BLEU4 | 0.67 | 0.41 | 0.67 |
| MedicationName Consideration | gpt-4 | ROUGE1 | 1.19 | 0.28 | 0.60 |
| MedicationName Consideration | gpt-4 | EM_F1 | 0.49 | 0.48 | 0.73 |
| Symptom Datetime | gpt-35-turbo | BLEU4 | 7.00 | 0.01 | 0.14 |
| Symptom Datetime | gpt-35-turbo | ROUGE1 | 12.73 | 0.00 | 0.07 |
| Symptom Datetime | gpt-35-turbo | EM_F1 | 9.60 | 0.00 | 0.08 |
| RadiologyTest Datetime | gpt-35-turbo | BLEU4 | 4.85 | 0.03 | 0.23 |
| RadiologyTest Datetime | gpt-35-turbo | ROUGE1 | 2.25 | 0.13 | 0.44 |
| RadiologyTest Datetime | gpt-35-turbo | EM_F1 | 0.43 | 0.51 | 0.75 |
| RadiologyTest Site | gpt-35-turbo | BLEU4 | 0.94 | 0.33 | 0.63 |

| Task | Model | Metric | Kruskal F-statistic | Kruskal p-value | Kruskal q-value |
|---|---|---|---|---|---|
| RadiologyTest Site | gpt-35-turbo | ROUGE1 | 0.08 | 0.78 | 0.93 |
| RadiologyTest Site | gpt-35-turbo | EM_F1 | 0.80 | 0.37 | 0.65 |
| RadiologyTest Reason | gpt-35-turbo | BLEU4 | 0.01 | 0.92 | 0.97 |
| RadiologyTest Reason | gpt-35-turbo | ROUGE1 | 0.07 | 0.80 | 0.93 |
| RadiologyTest Reason | gpt-35-turbo | EM_F1 | 0.01 | 0.94 | 0.97 |
| RadiologyTest Result | gpt-35-turbo | BLEU4 | 1.29 | 0.26 | 0.60 |
| RadiologyTest Result | gpt-35-turbo | ROUGE1 | 0.30 | 0.59 | 0.80 |
| RadiologyTest Result | gpt-35-turbo | EM_F1 | 0.05 | 0.83 | 0.93 |
| ProcedureName Datetime | gpt-35-turbo | BLEU4 | 0.01 | 0.90 | 0.96 |
| ProcedureName Datetime | gpt-35-turbo | ROUGE1 | 0.42 | 0.52 | 0.75 |
| ProcedureName Datetime | gpt-35-turbo | EM_F1 | 4.28 | 0.04 | 0.28 |
| ProcedureName Site | gpt-35-turbo | BLEU4 | 6.31 | 0.01 | 0.17 |
| ProcedureName Site | gpt-35-turbo | ROUGE1 | 2.06 | 0.15 | 0.46 |
| ProcedureName Site | gpt-35-turbo | EM_F1 | 3.30 | 0.07 | 0.34 |
| ProcedureName Reason | gpt-35-turbo | BLEU4 | 2.20 | 0.14 | 0.44 |
| ProcedureName Reason | gpt-35-turbo | ROUGE1 | 1.91 | 0.17 | 0.48 |
| ProcedureName Reason | gpt-35-turbo | EM_F1 | 0.21 | 0.64 | 0.83 |
| ProcedureName Result | gpt-35-turbo | BLEU4 | 1.52 | 0.22 | 0.57 |
| ProcedureName Result | gpt-35-turbo | ROUGE1 | 0.66 | 0.42 | 0.67 |
| ProcedureName Result | gpt-35-turbo | EM_F1 | 1.13 | 0.29 | 0.61 |
| GenomicTestName Datetime | gpt-35-turbo | BLEU4 | 3.08 | 0.08 | 0.34 |
| GenomicTestName Datetime | gpt-35-turbo | ROUGE1 | 3.06 | 0.08 | 0.34 |
| GenomicTestName Datetime | gpt-35-turbo | EM_F1 | 4.64 | 0.03 | 0.24 |
| GenomicTestName Result | gpt-35-turbo | BLEU4 | 1.53 | 0.22 | 0.57 |
| GenomicTestName Result | gpt-35-turbo | ROUGE1 | 3.63 | 0.06 | 0.34 |
| GenomicTestName Result | gpt-35-turbo | EM_F1 | 3.30 | 0.07 | 0.34 |
| Biomarker Datetime | gpt-35-turbo | BLEU4 | 7.37 | 0.01 | 0.13 |
| Biomarker Datetime | gpt-35-turbo | ROUGE1 | 11.67 | 0.00 | 0.07 |
| Biomarker Datetime | gpt-35-turbo | EM_F1 | 0.06 | 0.81 | 0.93 |

| Task | Model | Metric | Kruskal F-statistic | Kruskal p-value | Kruskal q-value |
| --- | --- | --- | --- | --- | --- |
| Histology Datetime | gpt-35-turbo | BLEU4 | 0.38 | 0.54 | 0.76 |
| Histology Datetime | gpt-35-turbo | ROUGE1 | 0.20 | 0.65 | 0.83 |
| Histology Datetime | gpt-35-turbo | EM_F1 | 0.21 | 0.64 | 0.83 |
| Metastasis Site | gpt-35-turbo | BLEU4 | 0.01 | 0.94 | 0.97 |
| Metastasis Site | gpt-35-turbo | ROUGE1 | 0.01 | 0.93 | 0.97 |
| Metastasis Site | gpt-35-turbo | EM_F1 | 0.50 | 0.48 | 0.73 |
| Metastasis Procedure | gpt-35-turbo | BLEU4 | 0.28 | 0.60 | 0.80 |
| Metastasis Procedure | gpt-35-turbo | ROUGE1 | 0.05 | 0.82 | 0.93 |
| Metastasis Procedure | gpt-35-turbo | EM_F1 | 0.41 | 0.52 | 0.75 |
| Metastasis Datetime | gpt-35-turbo | BLEU4 | 1.15 | 0.28 | 0.61 |
| Metastasis Datetime | gpt-35-turbo | ROUGE1 | 0.09 | 0.76 | 0.91 |
| Metastasis Datetime | gpt-35-turbo | EM_F1 | 0.46 | 0.50 | 0.74 |
| Stage Datetime | gpt-35-turbo | BLEU4 | 8.26 | 0.00 | 0.10 |
| Stage Datetime | gpt-35-turbo | ROUGE1 | 5.93 | 0.01 | 0.17 |
| Stage Datetime | gpt-35-turbo | EM_F1 | 9.63 | 0.00 | 0.08 |
| TNM Datetime | gpt-35-turbo | BLEU4 | 0.15 | 0.70 | 0.86 |
| TNM Datetime | gpt-35-turbo | ROUGE1 | 0.05 | 0.82 | 0.93 |
| Grade Datetime | gpt-35-turbo | BLEU4 | 5.06 | 0.02 | 0.21 |
| Grade Datetime | gpt-35-turbo | ROUGE1 | 7.49 | 0.01 | 0.13 |
| Grade Datetime | gpt-35-turbo | EM_F1 | 7.54 | 0.01 | 0.13 |
| MedicationName Begin | gpt-35-turbo | BLEU4 | 0.50 | 0.48 | 0.73 |
| MedicationName Begin | gpt-35-turbo | ROUGE1 | 0.05 | 0.82 | 0.93 |
| MedicationName Begin | gpt-35-turbo | EM_F1 | 0.55 | 0.46 | 0.72 |
| MedicationName End | gpt-35-turbo | BLEU4 | 1.69 | 0.19 | 0.52 |
| MedicationName End | gpt-35-turbo | ROUGE1 | 0.01 | 0.91 | 0.97 |
| MedicationName End | gpt-35-turbo | EM_F1 | 0.38 | 0.54 | 0.76 |
| MedicationName Reason | gpt-35-turbo | BLEU4 | 0.31 | 0.58 | 0.80 |
| MedicationName Reason | gpt-35-turbo | ROUGE1 | 0.33 | 0.56 | 0.79 |
| MedicationName Reason | gpt-35-turbo | EM_F1 | 0.74 | 0.39 | 0.67 |

| Task | Model | Metric | Kruskal F-statistic | Kruskal p-value | Kruskal q-value |
| --- | --- | --- | --- | --- | --- |
| MedicationName Continuity | gpt-35-turbo | BLEU4 | 0.29 | 0.59 | 0.80 |
| MedicationName Continuity | gpt-35-turbo | ROUGE1 | 0.64 | 0.42 | 0.67 |
| MedicationName Continuity | gpt-35-turbo | EM_F1 | 0.01 | 0.93 | 0.97 |
| MedicationName ConfirmedAdvEvent | gpt-35-turbo | BLEU4 | 0.01 | 0.92 | 0.97 |
| MedicationName ConfirmedAdvEvent | gpt-35-turbo | ROUGE1 | 0.42 | 0.52 | 0.75 |
| MedicationName ConfirmedAdvEvent | gpt-35-turbo | EM_F1 | 0.03 | 0.86 | 0.93 |
| MedicationName PotentialAdvEvent | gpt-35-turbo | BLEU4 | 1.18 | 0.28 | 0.60 |
| MedicationName PotentialAdvEvent | gpt-35-turbo | ROUGE1 | 5.91 | 0.02 | 0.17 |
| MedicationName PotentialAdvEvent | gpt-35-turbo | EM_F1 | 0.92 | 0.34 | 0.63 |
| MedicationName Consideration | gpt-35-turbo | BLEU4 | 0.00 | 0.97 | 0.99 |
| MedicationName Consideration | gpt-35-turbo | ROUGE1 | 1.33 | 0.25 | 0.60 |
| MedicationName Consideration | gpt-35-turbo | EM_F1 | 0.05 | 0.83 | 0.93 |
| Symptom Datetime | FLAN-UL2 | BLEU4 | 0.00 | 0.99 | 1.00 |
| Symptom Datetime | FLAN-UL2 | ROUGE1 | 0.28 | 0.60 | 0.80 |
| Symptom Datetime | FLAN-UL2 | EM_F1 | 0.26 | 0.61 | 0.81 |
| RadiologyTest Datetime | FLAN-UL2 | BLEU4 | 2.96 | 0.09 | 0.34 |
| RadiologyTest Datetime | FLAN-UL2 | ROUGE1 | 0.02 | 0.90 | 0.96 |
| RadiologyTest Datetime | FLAN-UL2 | EM_F1 | 0.07 | 0.79 | 0.93 |
| RadiologyTest Site | FLAN-UL2 | BLEU4 | 0.18 | 0.67 | 0.84 |
| RadiologyTest Site | FLAN-UL2 | ROUGE1 | 0.04 | 0.84 | 0.93 |
| RadiologyTest Reason | FLAN-UL2 | BLEU4 | 2.44 | 0.12 | 0.40 |
| RadiologyTest Reason | FLAN-UL2 | ROUGE1 | 0.00 | 0.97 | 0.99 |
| RadiologyTest Reason | FLAN-UL2 | EM_F1 | 0.00 | 1.00 | 1.00 |
| RadiologyTest Result | FLAN-UL2 | BLEU4 | 0.04 | 0.84 | 0.93 |
| RadiologyTest Result | FLAN-UL2 | ROUGE1 | 1.28 | 0.26 | 0.60 |
| RadiologyTest Result | FLAN-UL2 | EM_F1 | 0.27 | 0.60 | 0.80 |

| Task | Model | Metric | Kruskal F-statistic | Kruskal p-value | Kruskal q-value |
|---|---|---|---|---|---|
| ProcedureName Datetime | FLAN-UL2 | BLEU4 | 3.39 | 0.07 | 0.34 |
| ProcedureName Datetime | FLAN-UL2 | ROUGE1 | 0.85 | 0.36 | 0.64 |
| ProcedureName Datetime | FLAN-UL2 | EM_F1 | 1.82 | 0.18 | 0.50 |
| ProcedureName Site | FLAN-UL2 | BLEU4 | 1.34 | 0.25 | 0.60 |
| ProcedureName Site | FLAN-UL2 | ROUGE1 | 0.67 | 0.41 | 0.67 |
| ProcedureName Reason | FLAN-UL2 | BLEU4 | 0.00 | 0.99 | 1.00 |
| ProcedureName Reason | FLAN-UL2 | ROUGE1 | 1.04 | 0.31 | 0.61 |
| ProcedureName Reason | FLAN-UL2 | EM_F1 | 0.71 | 0.40 | 0.67 |
| ProcedureName Result | FLAN-UL2 | BLEU4 | 0.16 | 0.69 | 0.86 |
| ProcedureName Result | FLAN-UL2 | ROUGE1 | 1.97 | 0.16 | 0.47 |
| ProcedureName Result | FLAN-UL2 | EM_F1 | 0.68 | 0.41 | 0.67 |
| GenomicTestName Datetime | FLAN-UL2 | BLEU4 | 0.21 | 0.65 | 0.83 |
| GenomicTestName Datetime | FLAN-UL2 | ROUGE1 | 1.02 | 0.31 | 0.61 |
| GenomicTestName Datetime | FLAN-UL2 | EM_F1 | 0.71 | 0.40 | 0.67 |
| GenomicTestName Result | FLAN-UL2 | BLEU4 | 1.26 | 0.26 | 0.60 |
| GenomicTestName Result | FLAN-UL2 | ROUGE1 | 0.40 | 0.52 | 0.75 |
| Biomarker Datetime | FLAN-UL2 | BLEU4 | 0.85 | 0.36 | 0.64 |
| Biomarker Datetime | FLAN-UL2 | ROUGE1 | 0.42 | 0.51 | 0.75 |
| Biomarker Datetime | FLAN-UL2 | EM_F1 | 0.23 | 0.64 | 0.83 |
| Histology Datetime | FLAN-UL2 | BLEU4 | 0.71 | 0.40 | 0.67 |
| Histology Datetime | FLAN-UL2 | ROUGE1 | 0.97 | 0.33 | 0.62 |
| Histology Datetime | FLAN-UL2 | EM_F1 | 0.01 | 0.94 | 0.97 |
| Metastasis Site | FLAN-UL2 | BLEU4 | 0.19 | 0.67 | 0.83 |
| Metastasis Site | FLAN-UL2 | ROUGE1 | 2.90 | 0.09 | 0.34 |
| Metastasis Site | FLAN-UL2 | EM_F1 | 0.68 | 0.41 | 0.67 |
| Metastasis Procedure | FLAN-UL2 | BLEU4 | 0.80 | 0.37 | 0.65 |
| Metastasis Procedure | FLAN-UL2 | ROUGE1 | 1.04 | 0.31 | 0.61 |
| Metastasis Datetime | FLAN-UL2 | BLEU4 | 0.30 | 0.58 | 0.80 |
| Metastasis Datetime | FLAN-UL2 | ROUGE1 | 0.99 | 0.32 | 0.62 |

| Task | Model | Metric | Kruskal F-statistic | Kruskal p-value | Kruskal q-value |
|---|---|---|---|---|---|
| Stage Datetime | FLAN-UL2 | BLEU4 | 3.61 | 0.06 | 0.34 |
| Stage Datetime | FLAN-UL2 | ROUGE1 | 3.39 | 0.07 | 0.34 |
| Stage Datetime | FLAN-UL2 | EM_F1 | 4.64 | 0.03 | 0.24 |
| TNM Datetime | FLAN-UL2 | BLEU4 | 0.88 | 0.35 | 0.64 |
| TNM Datetime | FLAN-UL2 | ROUGE1 | 1.10 | 0.29 | 0.61 |
| TNM Datetime | FLAN-UL2 | EM_F1 | 0.45 | 0.50 | 0.75 |
| Grade Datetime | FLAN-UL2 | BLEU4 | 8.37 | 0.00 | 0.10 |
| Grade Datetime | FLAN-UL2 | ROUGE1 | 5.93 | 0.01 | 0.17 |
| Grade Datetime | FLAN-UL2 | EM_F1 | 2.92 | 0.09 | 0.34 |
| MedicationName Begin | FLAN-UL2 | BLEU4 | 3.88 | 0.05 | 0.33 |
| MedicationName Begin | FLAN-UL2 | ROUGE1 | 1.19 | 0.27 | 0.60 |
| MedicationName Begin | FLAN-UL2 | EM_F1 | 0.33 | 0.57 | 0.79 |
| MedicationName End | FLAN-UL2 | BLEU4 | 0.04 | 0.84 | 0.93 |
| MedicationName End | FLAN-UL2 | ROUGE1 | 3.20 | 0.07 | 0.34 |
| MedicationName End | FLAN-UL2 | EM_F1 | 0.64 | 0.42 | 0.67 |
| MedicationName Reason | FLAN-UL2 | BLEU4 | 6.23 | 0.01 | 0.17 |
| MedicationName Reason | FLAN-UL2 | ROUGE1 | 9.61 | 0.00 | 0.08 |
| MedicationName Reason | FLAN-UL2 | EM_F1 | 0.68 | 0.41 | 0.67 |
| MedicationName Continuity | FLAN-UL2 | BLEU4 | 3.04 | 0.08 | 0.34 |
| MedicationName Continuity | FLAN-UL2 | ROUGE1 | 3.07 | 0.08 | 0.34 |
| MedicationName Continuity | FLAN-UL2 | EM_F1 | 2.68 | 0.10 | 0.36 |
| MedicationName ConfirmedAdvEvent | FLAN-UL2 | BLEU4 | 0.54 | 0.46 | 0.72 |
| MedicationName ConfirmedAdvEvent | FLAN-UL2 | ROUGE1 | 2.91 | 0.09 | 0.34 |
| MedicationName ConfirmedAdvEvent | FLAN-UL2 | EM_F1 | 1.50 | 0.22 | 0.57 |
| MedicationName PotentialAdvEvent | FLAN-UL2 | BLEU4 | 0.12 | 0.73 | 0.90 |
| MedicationName PotentialAdvEvent | FLAN-UL2 | ROUGE1 | 0.12 | 0.73 | 0.89 |
| MedicationName PotentialAdvEvent | FLAN-UL2 | EM_F1 | 1.71 | 0.19 | 0.52 |

| Task | Model | Metric | Kruskal F-statistic | Kruskal p-value | Kruskal q-value |
| --- | --- | --- | --- | --- | --- |
| MedicationName Consideration | FLAN-UL2 | BLEU4 | 0.05 | 0.82 | 0.93 |
| MedicationName Consideration | FLAN-UL2 | ROUGE1 | 3.61 | 0.06 | 0.34 |
| MedicationName Consideration | FLAN-UL2 | EM_F1 | 0.09 | 0.76 | 0.91 |

Table S3: Kruskal-Wallis test statistics for comparing model performance across the included race/ethnicity variables for each information extraction task and metric. The q values are computed after correcting False Discovery Rate with Benjamini–Hochberg method.

| Task | Model | Metric | Kruskal F-statistic | Kruskal p-value | Kruskal q-value |
| --- | --- | --- | --- | --- | --- |
| Symptom Datetime | gpt-4 | BLEU4 | 12.64 | 0.12 | 0.79 |
| Symptom Datetime | gpt-4 | ROUGE1 | 5.39 | 0.72 | 0.93 |
| Symptom Datetime | gpt-4 | EM_F1 | 5.50 | 0.70 | 0.93 |
| RadiologyTest Datetime | gpt-4 | BLEU4 | 8.69 | 0.37 | 0.82 |
| RadiologyTest Datetime | gpt-4 | ROUGE1 | 9.27 | 0.32 | 0.82 |
| RadiologyTest Datetime | gpt-4 | EM_F1 | 8.78 | 0.36 | 0.82 |
| RadiologyTest Site | gpt-4 | BLEU4 | 3.39 | 0.91 | 0.96 |
| RadiologyTest Site | gpt-4 | ROUGE1 | 8.46 | 0.39 | 0.83 |
| RadiologyTest Site | gpt-4 | EM_F1 | 4.13 | 0.84 | 0.95 |
| RadiologyTest Reason | gpt-4 | BLEU4 | 5.96 | 0.65 | 0.91 |
| RadiologyTest Reason | gpt-4 | ROUGE1 | 6.24 | 0.62 | 0.90 |
| RadiologyTest Reason | gpt-4 | EM_F1 | 6.27 | 0.62 | 0.90 |
| RadiologyTest Result | gpt-4 | BLEU4 | 6.34 | 0.61 | 0.90 |
| RadiologyTest Result | gpt-4 | ROUGE1 | 5.84 | 0.67 | 0.92 |
| RadiologyTest Result | gpt-4 | EM_F1 | 5.22 | 0.73 | 0.93 |
| ProcedureName Datetime | gpt-4 | BLEU4 | 2.17 | 0.98 | 0.98 |
| ProcedureName Datetime | gpt-4 | ROUGE1 | 4.35 | 0.82 | 0.95 |
| ProcedureName Datetime | gpt-4 | EM_F1 | 5.10 | 0.75 | 0.93 |
| ProcedureName Site | gpt-4 | BLEU4 | 3.32 | 0.91 | 0.96 |

| Task | Model | Metric | Kruskal F-statistic | Kruskal p-value | Kruskal q-value |
|---|---|---|---|---|---|
| ProcedureName Site | gpt-4 | ROUGE1 | 3.42 | 0.91 | 0.96 |
| ProcedureName Site | gpt-4 | EM_F1 | 4.68 | 0.79 | 0.95 |
| ProcedureName Reason | gpt-4 | BLEU4 | 3.18 | 0.92 | 0.96 |
| ProcedureName Reason | gpt-4 | ROUGE1 | 4.00 | 0.86 | 0.95 |
| ProcedureName Reason | gpt-4 | EM_F1 | 5.16 | 0.74 | 0.93 |
| ProcedureName Result | gpt-4 | BLEU4 | 1.98 | 0.98 | 0.99 |
| ProcedureName Result | gpt-4 | ROUGE1 | 3.36 | 0.91 | 0.96 |
| ProcedureName Result | gpt-4 | EM_F1 | 7.29 | 0.51 | 0.90 |
| GenomicTestName Datetime | gpt-4 | BLEU4 | 9.58 | 0.30 | 0.79 |
| GenomicTestName Datetime | gpt-4 | ROUGE1 | 11.10 | 0.20 | 0.79 |
| GenomicTestName Datetime | gpt-4 | EM_F1 | 12.63 | 0.13 | 0.79 |
| GenomicTestName Result | gpt-4 | BLEU4 | 10.11 | 0.26 | 0.79 |
| GenomicTestName Result | gpt-4 | ROUGE1 | 10.47 | 0.23 | 0.79 |
| GenomicTestName Result | gpt-4 | EM_F1 | 13.54 | 0.09 | 0.79 |
| Biomarker Datetime | gpt-4 | BLEU4 | 11.70 | 0.17 | 0.79 |
| Biomarker Datetime | gpt-4 | ROUGE1 | 8.84 | 0.36 | 0.82 |
| Biomarker Datetime | gpt-4 | EM_F1 | 8.17 | 0.42 | 0.87 |
| Histology Datetime | gpt-4 | BLEU4 | 11.11 | 0.20 | 0.79 |
| Histology Datetime | gpt-4 | ROUGE1 | 7.67 | 0.47 | 0.90 |
| Histology Datetime | gpt-4 | EM_F1 | 10.56 | 0.23 | 0.79 |
| Metastasis Site | gpt-4 | BLEU4 | 11.71 | 0.16 | 0.79 |
| Metastasis Site | gpt-4 | ROUGE1 | 7.53 | 0.48 | 0.90 |
| Metastasis Site | gpt-4 | EM_F1 | 12.40 | 0.13 | 0.79 |
| Metastasis Procedure | gpt-4 | BLEU4 | 8.39 | 0.40 | 0.83 |
| Metastasis Procedure | gpt-4 | ROUGE1 | 6.32 | 0.61 | 0.90 |
| Metastasis Procedure | gpt-4 | EM_F1 | 11.21 | 0.19 | 0.79 |
| Metastasis Datetime | gpt-4 | BLEU4 | 9.18 | 0.33 | 0.82 |
| Metastasis Datetime | gpt-4 | ROUGE1 | 6.51 | 0.59 | 0.90 |
| Metastasis Datetime | gpt-4 | EM_F1 | 11.10 | 0.20 | 0.79 |

| Task | Model | Metric | Kruskal F-statistic | Kruskal p-value | Kruskal q-value |
| --- | --- | --- | --- | --- | --- |
| Stage Datetime | gpt-4 | BLEU4 | 11.28 | 0.19 | 0.79 |
| Stage Datetime | gpt-4 | ROUGE1 | 12.25 | 0.14 | 0.79 |
| Stage Datetime | gpt-4 | EM_F1 | 12.31 | 0.14 | 0.79 |
| TNM Datetime | gpt-4 | BLEU4 | 12.67 | 0.12 | 0.79 |
| TNM Datetime | gpt-4 | ROUGE1 | 12.51 | 0.13 | 0.79 |
| Grade Datetime | gpt-4 | BLEU4 | 6.18 | 0.63 | 0.91 |
| Grade Datetime | gpt-4 | ROUGE1 | 4.24 | 0.83 | 0.95 |
| Grade Datetime | gpt-4 | EM_F1 | 4.37 | 0.82 | 0.95 |
| MedicationName Begin | gpt-4 | BLEU4 | 8.50 | 0.39 | 0.83 |
| MedicationName Begin | gpt-4 | ROUGE1 | 5.45 | 0.71 | 0.93 |
| MedicationName Begin | gpt-4 | EM_F1 | 7.34 | 0.50 | 0.90 |
| MedicationName End | gpt-4 | BLEU4 | 10.34 | 0.24 | 0.79 |
| MedicationName End | gpt-4 | ROUGE1 | 7.00 | 0.54 | 0.90 |
| MedicationName End | gpt-4 | EM_F1 | 9.08 | 0.34 | 0.82 |
| MedicationName Reason | gpt-4 | BLEU4 | 9.81 | 0.28 | 0.79 |
| MedicationName Reason | gpt-4 | ROUGE1 | 8.74 | 0.36 | 0.82 |
| MedicationName Reason | gpt-4 | EM_F1 | 11.26 | 0.19 | 0.79 |
| MedicationName Continuity | gpt-4 | BLEU4 | 10.03 | 0.26 | 0.79 |
| MedicationName Continuity | gpt-4 | ROUGE1 | 7.64 | 0.47 | 0.90 |
| MedicationName Continuity | gpt-4 | EM_F1 | 9.49 | 0.30 | 0.79 |
| MedicationName ConfirmedAdvEvent | gpt-4 | BLEU4 | 10.75 | 0.22 | 0.79 |
| MedicationName ConfirmedAdvEvent | gpt-4 | ROUGE1 | 9.75 | 0.28 | 0.79 |
| MedicationName ConfirmedAdvEvent | gpt-4 | EM_F1 | 6.54 | 0.59 | 0.90 |
| MedicationName PotentialAdvEvent | gpt-4 | BLEU4 | 9.21 | 0.32 | 0.82 |
| MedicationName PotentialAdvEvent | gpt-4 | ROUGE1 | 5.42 | 0.71 | 0.93 |
| MedicationName PotentialAdvEvent | gpt-4 | EM_F1 | 6.75 | 0.56 | 0.90 |
| MedicationName Consideration | gpt-4 | BLEU4 | 8.49 | 0.39 | 0.83 |
| MedicationName Consideration | gpt-4 | ROUGE1 | 5.46 | 0.71 | 0.93 |
| MedicationName Consideration | gpt-4 | EM_F1 | 5.43 | 0.71 | 0.93 |

| Task | Model | Metric | Kruskal F-statistic | Kruskal p-value | Kruskal q-value |
|---|---|---|---|---|---|
| Symptom Datetime | gpt-35-turbo | BLEU4 | 11.02 | 0.20 | 0.79 |
| Symptom Datetime | gpt-35-turbo | ROUGE1 | 6.50 | 0.59 | 0.90 |
| Symptom Datetime | gpt-35-turbo | EM_F1 | 8.83 | 0.36 | 0.82 |
| RadiologyTest Datetime | gpt-35-turbo | BLEU4 | 8.83 | 0.36 | 0.82 |
| RadiologyTest Datetime | gpt-35-turbo | ROUGE1 | 10.91 | 0.21 | 0.79 |
| RadiologyTest Datetime | gpt-35-turbo | EM_F1 | 11.59 | 0.17 | 0.79 |
| RadiologyTest Site | gpt-35-turbo | BLEU4 | 6.08 | 0.64 | 0.91 |
| RadiologyTest Site | gpt-35-turbo | ROUGE1 | 12.48 | 0.13 | 0.79 |
| RadiologyTest Site | gpt-35-turbo | EM_F1 | 7.73 | 0.46 | 0.90 |
| RadiologyTest Reason | gpt-35-turbo | BLEU4 | 2.67 | 0.95 | 0.97 |
| RadiologyTest Reason | gpt-35-turbo | ROUGE1 | 7.66 | 0.47 | 0.90 |
| RadiologyTest Reason | gpt-35-turbo | EM_F1 | 10.84 | 0.21 | 0.79 |
| RadiologyTest Result | gpt-35-turbo | BLEU4 | 7.19 | 0.52 | 0.90 |
| RadiologyTest Result | gpt-35-turbo | ROUGE1 | 10.63 | 0.22 | 0.79 |
| RadiologyTest Result | gpt-35-turbo | EM_F1 | 9.03 | 0.34 | 0.82 |
| ProcedureName Datetime | gpt-35-turbo | BLEU4 | 10.37 | 0.24 | 0.79 |
| ProcedureName Datetime | gpt-35-turbo | ROUGE1 | 15.32 | 0.05 | 0.63 |
| ProcedureName Datetime | gpt-35-turbo | EM_F1 | 4.28 | 0.83 | 0.95 |
| ProcedureName Site | gpt-35-turbo | BLEU4 | 1.17 | 1.00 | 1.00 |
| ProcedureName Site | gpt-35-turbo | ROUGE1 | 4.08 | 0.85 | 0.95 |
| ProcedureName Site | gpt-35-turbo | EM_F1 | 6.40 | 0.60 | 0.90 |
| ProcedureName Reason | gpt-35-turbo | BLEU4 | 6.68 | 0.57 | 0.90 |
| ProcedureName Reason | gpt-35-turbo | ROUGE1 | 8.09 | 0.42 | 0.88 |
| ProcedureName Reason | gpt-35-turbo | EM_F1 | 9.92 | 0.27 | 0.79 |
| ProcedureName Result | gpt-35-turbo | BLEU4 | 7.08 | 0.53 | 0.90 |
| ProcedureName Result | gpt-35-turbo | ROUGE1 | 6.28 | 0.62 | 0.90 |
| ProcedureName Result | gpt-35-turbo | EM_F1 | 9.01 | 0.34 | 0.82 |
| GenomicTestName Datetime | gpt-35-turbo | BLEU4 | 6.36 | 0.61 | 0.90 |
| GenomicTestName Datetime | gpt-35-turbo | ROUGE1 | 10.54 | 0.23 | 0.79 |

| Task | Model | Metric | Kruskal F-statistic | Kruskal p-value | Kruskal q-value |
|---|---|---|---|---|---|
| GenomicTestName Datetime | gpt-35-turbo | EM_F1 | 7.88 | 0.45 | 0.90 |
| GenomicTestName Result | gpt-35-turbo | BLEU4 | 2.63 | 0.96 | 0.97 |
| GenomicTestName Result | gpt-35-turbo | ROUGE1 | 5.96 | 0.65 | 0.91 |
| GenomicTestName Result | gpt-35-turbo | EM_F1 | 4.23 | 0.84 | 0.95 |
| Biomarker Datetime | gpt-35-turbo | BLEU4 | 14.84 | 0.06 | 0.67 |
| Biomarker Datetime | gpt-35-turbo | ROUGE1 | 16.05 | 0.04 | 0.56 |
| Biomarker Datetime | gpt-35-turbo | EM_F1 | 7.28 | 0.51 | 0.90 |
| Histology Datetime | gpt-35-turbo | BLEU4 | 7.55 | 0.48 | 0.90 |
| Histology Datetime | gpt-35-turbo | ROUGE1 | 5.04 | 0.75 | 0.93 |
| Histology Datetime | gpt-35-turbo | EM_F1 | 6.79 | 0.56 | 0.90 |
| Metastasis Site | gpt-35-turbo | BLEU4 | 9.11 | 0.33 | 0.82 |
| Metastasis Site | gpt-35-turbo | ROUGE1 | 8.95 | 0.35 | 0.82 |
| Metastasis Site | gpt-35-turbo | EM_F1 | 12.65 | 0.12 | 0.79 |
| Metastasis Procedure | gpt-35-turbo | BLEU4 | 6.42 | 0.60 | 0.90 |
| Metastasis Procedure | gpt-35-turbo | ROUGE1 | 10.21 | 0.25 | 0.79 |
| Metastasis Procedure | gpt-35-turbo | EM_F1 | 13.73 | 0.09 | 0.79 |
| Metastasis Datetime | gpt-35-turbo | BLEU4 | 9.58 | 0.30 | 0.79 |
| Metastasis Datetime | gpt-35-turbo | ROUGE1 | 9.79 | 0.28 | 0.79 |
| Metastasis Datetime | gpt-35-turbo | EM_F1 | 13.69 | 0.09 | 0.79 |
| Stage Datetime | gpt-35-turbo | BLEU4 | 3.29 | 0.91 | 0.96 |
| Stage Datetime | gpt-35-turbo | ROUGE1 | 2.56 | 0.96 | 0.97 |
| Stage Datetime | gpt-35-turbo | EM_F1 | 4.48 | 0.81 | 0.95 |
| TNM Datetime | gpt-35-turbo | BLEU4 | 6.91 | 0.55 | 0.90 |
| TNM Datetime | gpt-35-turbo | ROUGE1 | 7.07 | 0.53 | 0.90 |
| Grade Datetime | gpt-35-turbo | BLEU4 | 21.01 | 0.01 | 0.20 |
| Grade Datetime | gpt-35-turbo | ROUGE1 | 16.51 | 0.04 | 0.54 |
| Grade Datetime | gpt-35-turbo | EM_F1 | 17.44 | 0.03 | 0.52 |
| MedicationName Begin | gpt-35-turbo | BLEU4 | 7.19 | 0.52 | 0.90 |
| MedicationName Begin | gpt-35-turbo | ROUGE1 | 6.87 | 0.55 | 0.90 |

| Task | Model | Metric | Kruskal F-statistic | Kruskal p-value | Kruskal q-value |
| --- | --- | --- | --- | --- | --- |
| MedicationName Begin | gpt-35-turbo | EM_F1 | 5.23 | 0.73 | 0.93 |
| MedicationName End | gpt-35-turbo | BLEU4 | 4.43 | 0.82 | 0.95 |
| MedicationName End | gpt-35-turbo | ROUGE1 | 3.63 | 0.89 | 0.96 |
| MedicationName End | gpt-35-turbo | EM_F1 | 3.01 | 0.93 | 0.96 |
| MedicationName Reason | gpt-35-turbo | BLEU4 | 6.93 | 0.54 | 0.90 |
| MedicationName Reason | gpt-35-turbo | ROUGE1 | 4.90 | 0.77 | 0.93 |
| MedicationName Reason | gpt-35-turbo | EM_F1 | 9.71 | 0.29 | 0.79 |
| MedicationName Continuity | gpt-35-turbo | BLEU4 | 3.05 | 0.93 | 0.96 |
| MedicationName Continuity | gpt-35-turbo | ROUGE1 | 3.56 | 0.89 | 0.96 |
| MedicationName Continuity | gpt-35-turbo | EM_F1 | 3.27 | 0.92 | 0.96 |
| MedicationName ConfirmedAdvEvent | gpt-35-turbo | BLEU4 | 3.91 | 0.87 | 0.96 |
| MedicationName ConfirmedAdvEvent | gpt-35-turbo | ROUGE1 | 4.91 | 0.77 | 0.93 |
| MedicationName ConfirmedAdvEvent | gpt-35-turbo | EM_F1 | 4.09 | 0.85 | 0.95 |
| MedicationName PotentialAdvEvent | gpt-35-turbo | BLEU4 | 5.74 | 0.68 | 0.93 |
| MedicationName PotentialAdvEvent | gpt-35-turbo | ROUGE1 | 3.64 | 0.89 | 0.96 |
| MedicationName PotentialAdvEvent | gpt-35-turbo | EM_F1 | 3.05 | 0.93 | 0.96 |
| MedicationName Consideration | gpt-35-turbo | BLEU4 | 6.37 | 0.61 | 0.90 |
| MedicationName Consideration | gpt-35-turbo | ROUGE1 | 2.97 | 0.94 | 0.96 |
| MedicationName Consideration | gpt-35-turbo | EM_F1 | 7.30 | 0.50 | 0.90 |
| Symptom Datetime | FLAN-UL2 | BLEU4 | 8.43 | 0.39 | 0.83 |
| Symptom Datetime | FLAN-UL2 | ROUGE1 | 6.96 | 0.54 | 0.90 |
| Symptom Datetime | FLAN-UL2 | EM_F1 | 22.64 | 0.00 | 0.16 |
| RadiologyTest Datetime | FLAN-UL2 | BLEU4 | 7.42 | 0.49 | 0.90 |
| RadiologyTest Datetime | FLAN-UL2 | ROUGE1 | 5.16 | 0.74 | 0.93 |
| RadiologyTest Datetime | FLAN-UL2 | EM_F1 | 5.61 | 0.69 | 0.93 |
| RadiologyTest Site | FLAN-UL2 | BLEU4 | 5.31 | 0.72 | 0.93 |
| RadiologyTest Site | FLAN-UL2 | ROUGE1 | 4.16 | 0.84 | 0.95 |
| RadiologyTest Reason | FLAN-UL2 | BLEU4 | 5.12 | 0.75 | 0.93 |
| RadiologyTest Reason | FLAN-UL2 | ROUGE1 | 4.38 | 0.82 | 0.95 |

| Task | Model | Metric | Kruskal F-statistic | Kruskal p-value | Kruskal q-value |
| --- | --- | --- | --- | --- | --- |
| RadiologyTest Reason | FLAN-UL2 | EM_F1 | 6.24 | 0.62 | 0.90 |
| RadiologyTest Result | FLAN-UL2 | BLEU4 | 6.62 | 0.58 | 0.90 |
| RadiologyTest Result | FLAN-UL2 | ROUGE1 | 3.02 | 0.93 | 0.96 |
| RadiologyTest Result | FLAN-UL2 | EM_F1 | 7.97 | 0.44 | 0.89 |
| ProcedureName Datetime | FLAN-UL2 | BLEU4 | 5.09 | 0.75 | 0.93 |
| ProcedureName Datetime | FLAN-UL2 | ROUGE1 | 6.00 | 0.65 | 0.91 |
| ProcedureName Datetime | FLAN-UL2 | EM_F1 | 11.28 | 0.19 | 0.79 |
| ProcedureName Site | FLAN-UL2 | BLEU4 | 5.27 | 0.73 | 0.93 |
| ProcedureName Site | FLAN-UL2 | ROUGE1 | 9.76 | 0.28 | 0.79 |
| ProcedureName Reason | FLAN-UL2 | BLEU4 | 6.62 | 0.58 | 0.90 |
| ProcedureName Reason | FLAN-UL2 | ROUGE1 | 11.77 | 0.16 | 0.79 |
| ProcedureName Reason | FLAN-UL2 | EM_F1 | 7.36 | 0.50 | 0.90 |
| ProcedureName Result | FLAN-UL2 | BLEU4 | 7.68 | 0.46 | 0.90 |
| ProcedureName Result | FLAN-UL2 | ROUGE1 | 13.36 | 0.10 | 0.79 |
| ProcedureName Result | FLAN-UL2 | EM_F1 | 3.04 | 0.93 | 0.96 |
| GenomicTestName Datetime | FLAN-UL2 | BLEU4 | 9.15 | 0.33 | 0.82 |
| GenomicTestName Datetime | FLAN-UL2 | ROUGE1 | 9.63 | 0.29 | 0.79 |
| GenomicTestName Datetime | FLAN-UL2 | EM_F1 | 11.48 | 0.18 | 0.79 |
| GenomicTestName Result | FLAN-UL2 | BLEU4 | 6.56 | 0.58 | 0.90 |
| GenomicTestName Result | FLAN-UL2 | ROUGE1 | 17.19 | 0.03 | 0.52 |
| Biomarker Datetime | FLAN-UL2 | BLEU4 | 9.48 | 0.30 | 0.79 |
| Biomarker Datetime | FLAN-UL2 | ROUGE1 | 4.46 | 0.81 | 0.95 |
| Biomarker Datetime | FLAN-UL2 | EM_F1 | 4.27 | 0.83 | 0.95 |
| Histology Datetime | FLAN-UL2 | BLEU4 | 10.11 | 0.26 | 0.79 |
| Histology Datetime | FLAN-UL2 | ROUGE1 | 16.34 | 0.04 | 0.54 |
| Histology Datetime | FLAN-UL2 | EM_F1 | 10.29 | 0.25 | 0.79 |
| Metastasis Site | FLAN-UL2 | BLEU4 | 15.10 | 0.06 | 0.65 |
| Metastasis Site | FLAN-UL2 | ROUGE1 | 10.28 | 0.25 | 0.79 |
| Metastasis Site | FLAN-UL2 | EM_F1 | 22.03 | 0.00 | 0.16 |

| Task | Model | Metric | Kruskal F-statistic | Kruskal p-value | Kruskal q-value |
|---|---|---|---|---|---|
| Metastasis Procedure | FLAN-UL2 | BLEU4 | 14.41 | 0.07 | 0.74 |
| Metastasis Procedure | FLAN-UL2 | ROUGE1 | 9.51 | 0.30 | 0.79 |
| Metastasis Datetime | FLAN-UL2 | BLEU4 | 5.93 | 0.65 | 0.91 |
| Metastasis Datetime | FLAN-UL2 | ROUGE1 | 8.58 | 0.38 | 0.83 |
| Stage Datetime | FLAN-UL2 | BLEU4 | 10.77 | 0.22 | 0.79 |
| Stage Datetime | FLAN-UL2 | ROUGE1 | 10.53 | 0.23 | 0.79 |
| Stage Datetime | FLAN-UL2 | EM_F1 | 19.54 | 0.01 | 0.31 |
| TNM Datetime | FLAN-UL2 | BLEU4 | 7.82 | 0.45 | 0.90 |
| TNM Datetime | FLAN-UL2 | ROUGE1 | 11.02 | 0.20 | 0.79 |
| TNM Datetime | FLAN-UL2 | EM_F1 | 11.82 | 0.16 | 0.79 |
| Grade Datetime | FLAN-UL2 | BLEU4 | 22.42 | 0.00 | 0.16 |
| Grade Datetime | FLAN-UL2 | ROUGE1 | 16.80 | 0.03 | 0.52 |
| Grade Datetime | FLAN-UL2 | EM_F1 | 13.43 | 0.10 | 0.79 |
| MedicationName Begin | FLAN-UL2 | BLEU4 | 28.19 | 0.00 | 0.06 |
| MedicationName Begin | FLAN-UL2 | ROUGE1 | 23.50 | 0.00 | 0.16 |
| MedicationName Begin | FLAN-UL2 | EM_F1 | 27.64 | 0.00 | 0.06 |
| MedicationName End | FLAN-UL2 | BLEU4 | 9.74 | 0.28 | 0.79 |
| MedicationName End | FLAN-UL2 | ROUGE1 | 17.28 | 0.03 | 0.52 |
| MedicationName End | FLAN-UL2 | EM_F1 | 22.58 | 0.00 | 0.16 |
| MedicationName Reason | FLAN-UL2 | BLEU4 | 11.29 | 0.19 | 0.79 |
| MedicationName Reason | FLAN-UL2 | ROUGE1 | 15.76 | 0.05 | 0.58 |
| MedicationName Reason | FLAN-UL2 | EM_F1 | 10.45 | 0.24 | 0.79 |
| MedicationName Continuity | FLAN-UL2 | BLEU4 | 10.77 | 0.22 | 0.79 |
| MedicationName Continuity | FLAN-UL2 | ROUGE1 | 17.06 | 0.03 | 0.52 |
| MedicationName Continuity | FLAN-UL2 | EM_F1 | 13.09 | 0.11 | 0.79 |
| MedicationName ConfirmedAdvEvent | FLAN-UL2 | BLEU4 | 12.62 | 0.13 | 0.79 |
| MedicationName ConfirmedAdvEvent | FLAN-UL2 | ROUGE1 | 7.65 | 0.47 | 0.90 |
| MedicationName ConfirmedAdvEvent | FLAN-UL2 | EM_F1 | 7.41 | 0.49 | 0.90 |
| MedicationName PotentialAdvEvent | FLAN-UL2 | BLEU4 | 4.99 | 0.76 | 0.93 |

| Task | Model | Metric | Kruskal F-statistic | Kruskal p-value | Kruskal q-value |
|---|---|---|---|---|---|
| MedicationName PotentialAdvEvent | FLAN-UL2 | ROUGE1 | 4.77 | 0.78 | 0.94 |
| MedicationName PotentialAdvEvent | FLAN-UL2 | EM_F1 | 11.94 | 0.15 | 0.79 |
| MedicationName Consideration | FLAN-UL2 | BLEU4 | 5.39 | 0.71 | 0.93 |
| MedicationName Consideration | FLAN-UL2 | ROUGE1 | 5.98 | 0.65 | 0.91 |
| MedicationName Consideration | FLAN-UL2 | EM_F1 | 5.56 | 0.70 | 0.93 |

## GPT-3.5-turbo and GPT-4 settings

0613 version of the GPT-3.5-turbo model and 0314 version of the GPT-4 model were used via the Microsoft Azure OpenAI studio platform for all the experiments. The API version was *2023-05-15*. The most deterministic temperature setting of 0 was used. The outputs were retrieved via the ChatCompletion API using the prompts described next. Additional prompt engineering or hyperparameter tuning was not performed.

## Prompts for extracting related pairs of entities for all the sub-tasks

**System role:**

*Pretend you are an oncologist. Answer based on the given clinical note for a patient.*

**User prompt template:**

*{task-specific-prompt}*

*Answer as concisely as possible.*

*Use '\' for special quotation characters.*

*Do not return any information not present in the note.*

*Do not return any explanation.*

**Task-specific user prompts:**

1. Symptom history

*For this note, please return all symptoms experienced by patient, paired with the date or time they experienced that symptom.*

*Do not return any medical diagnoses, radiological findings, clinical test, or procedure results.*

*If a symptom is discussed, but only as a potential side effect or in the context of confirming the symptom's absence, please do not include this symptom.*

*In addition to returning the symptoms, return the date or time of the symptom onset.*

*If the date or time is not present within the note, please return 'unknown' in the given format.*

*Please return as namedtuples separated by newlines in the following format:*

*SymptomEnt(Symptom='Symptom identified', Datetime={'Date or time identified'})*

*Example:*

*SymptomEnt(Symptom='Abdominal Pain', Datetime={'01/01/2020', '02/03/2020'})*

*SymptomEnt(Symptom='lump', Datetime={'unknown'})*

2. Symptoms present at time of cancer diagnosis

    *First, identify the date of first cancer diagnosis for this patient.*

    *After you have done this, please return symptoms experienced by the patient that were present before or at the time of cancer diagnosis.*

    *If present, pair these with the date or time they started experiencing that symptom.*

    *If the date or time is not present within the note, please return 'unknown' in the given format.*

    *Do not return any medical diagnoses, radiological findings, clinical test or procedure results.*

    *Please return as namedtuples separated by newlines in the following format:*

    *CancerDiagnosis(Datetime={'Date or time of Cancer Diagnosis'})*

    *SymptomEnt(Symptom='Symptom identified', Datetime={'Date or time identified'})*

    *One example:*

    *CancerDiagnosis(Datetime={'03/01/2019'})*

    *SymptomEnt(Symptom='Abdominal Pain', Datetime={'01/01/2019'})*

    *Do not return any symptoms patient started experiencing subsequent to cancer diagnosis.*

3. Symptoms likely to be caused by the cancer

*For this note, please return all symptoms experienced by patient LIKELY TO BE CAUSED BY CANCER, paired with the date or time they started experiencing that symptom.*

*If the date or time is not present within the note, please return 'unknown' in the given format.*

*Do not return any medical diagnoses, radiological findings, clinical test or procedure results.*

*Please return as namedtuples separated by newlines in the following format:*

*SymptomEnt(Symptom='Symptom identified'', Datetime={'Date or time identified'})*

*Example:*

*SymptomEnt(Symptom='Abdominal Pain', Datetime={'01/01/2020', '02/03/2020'})*

*SymptomEnt(Symptom='lump', Datetime={'unknown'})*

4. Radiology tests

*For this note, please return all Radiology studies conducted for the patient, paired with the date or time when the study was performed, site of the study with its laterality, the symptom or clinical finding that the test was conducted for, and the results.*

*Only include the results that are of relevance to an Oncologist.*

*If any information is not present within the note, please return 'unknown'.*

*Please return as namedtuples separated by newlines in the following format:*

*RadTest(RadiologyTest='Radiology test', Datetime={'Date or time'}, Site={'Laterality and Site of test'}, Reason={'Symptom or clinical finding for which the test was conducted'}, Result={'Result of the test'})*

*Example:*

*RadTest(RadiologyTest='MRI', Datetime={'01/01/2020'}, Site={'left breast'}, Reason={'lump'}, Result={'abnormal mass'})*

*RadTest(RadiologyTest='PETCT', Datetime={'unknown'}, Site={'left, right breast'}, Reason={'unknown'}, Result={'unknown'})*

5. Medical procedures

   *For this note, please return all cancer-directed diagnostic and interventional procedures where there is a risk for bleeding.*
   *Pair these procedures with the date or time that the procedure was performed, laterality and site of the procedure, the clinical condition (e.g. diagnosis, symptom or problem) that the procedure was meant to identify or treat, and the result of the procedure.*
   *Only include the results that are of relevance to an Oncologist.*
   *If the information is not present within the note, please return 'unknown'.*
   *Please return as namedtuples separated by newlines in the following format:*
   *Proc(ProcedureName='Procedure identified', Datetime={'Date or time'}, Site={'laterality and site of procedure'}, Reason={'the clinical condition, such as diagnosis, symptom or problem that the procedure was meant to identify or treat'}, Result={'Result of the procedure'})*
   *Example:*
   *Proc(ProcedureName='Partial mastectomy', Datetime={'unknown'}, Site={'right breast'}, Reason={'lump'}, Result={'invasive carcinoma'})*
   *Proc(ProcedureName='mastectomy', Datetime={'01/01/2020'}, Site={'left, right breast'}, Reason={'unknown'}, Result={'unknown'})*
   *DO NOT return radiology tests.*

6. Genomic tests

   *For this note, please return all genomic and genetic tests that were conducted for the patient, and pair it with the date or time of the test and the result of the test.*
   *If any information is not present within the note, please return 'unknown'.*
   *Please return as namedtuples separated by newlines in the following format:*
   *Genomics(GenomicTestName='Name of genomic test conducted', Datetime={'Date or time'}, Result={'Result of the test'})*
   *Example:*
   *Genomics(GenomicTestName='Onctotype', Datetime={'unknown'}, Result={'high risk'})*

*Genomics(GenomicTestName='Genetic panel testing', Datetime={'01/01/2021'}, Result={'unknown'})*

*Do not return radiology tests, surgical procedures, or cancer biomarkers that do not have a genomic or genetic basis.*

7. Tumor Characteristics

    a. Treatment relevant biomarkers

*For this note, please return all treatment relevant biomarkers identified for the main cancer of this patient, paired with the date or time for that the characteristic was identified.*
*If the date that identified the characteristic is not present within the note, please return 'unknown'.*
*Please return as namedtuples separated by newlines in the following format:*
*TxBiomarker(Biomarker='biomarker name and result", Datetime={'Date or time identified'})*
*Example:*
*TxBiomarker(Biomarker='ER+', Datetime={'01/01/2020', '03/01/2021'})*
*TxBiomarker(Biomarker='HER2-', Datetime={'unknown'})*
*Do NOT return any radiological findings.*

    b. Specific histopathology subtype

*For this note, please return all morphological histology types identified for main cancer of the patient, paired with the date that characteristic was identified.*
*If the date for the identified characteristic is not present within the note, please return 'unknown'.*
*Please return as namedtuples separated by newlines in the following format:*
*Histo(Histology='Histopathology', Datetime={'Date or time identified'})*
*Example:*
*Histo(Histology='IDC', Datetime={'unknown'})*
*Histo(Histology='invasive ductal carcinoma', Datetime={'01/01/2020', '02/01/2013'})*
*Do NOT return any radiological findings.*

    c. Evidence of Metastatic spread

*For this note, please return any evidence of metastatic spread identified for the cancer of this patient, paired with the procedure and date that identified the metastatic spread.*

*If any information about the metastatic spread is not present within the note, please return 'unknown'.*

*Please return as namedtuples separated by newlines in the following format:*

*MetastasisEnt(Metastasis='mention of Metastatic spread', Site={'Site of spread'}, Procedure={'Procedure that identified the metastatic spread'}, Datetime={'Date or time the procedure that identified the spread was conducted'})*

*Example:*

*MetastasisEnt(Metastasis='metastasis', Procedure={'CT Chest'}, Datetime={'01/01/2020'}, Site={'lungs'})*

*MetastasisEnt(Metastasis='metastasis', Procedure={'unknown'}, Datetime={'unknown'}, Site={'nodes'})*

    d. *Stage of patient*

*For this note, please return staging for the main cancer of this patient, paired with the date on which this staging was done.*

*Do not include any stage indicated by the TNM staging criteria.*

*If additional testing must be done to fully stage the patient, please return this within the same format as described below or return unclear.*

*If the date of first staging is not clear in note, please return 'unknown'.*

*Please return as namedtuples separated by newlines in the following format:*

*StageEnt(Stage='Stage', Datetime={'Date conducted'}, AdditionalTesting={'test to be done'})*

*Example:*

*StageEnt(Stage='II', Datetime={'01/01/2020'}, AdditionalTesting={'CT Chest, abdomen, Pelvis'})*

*StageEnt(Stage='early', Datetime={'unknown'}, AdditionalTesting={'unknown'})*

    e. *Stage of patient (TNM)*

*For this note, please return TNM Staging system for main cancer.*

*If additional testing must be done to fully stage the patient by TNM staging, please return this within the same format as described below.*

*If the date of first staging is not clear in note, please return 'unknown'.*

*Please return as namedtuples separated by newlines in the following format:*

*TnmEnt(TNM='TNM Stage', Datetime={'Date conducted'}, AdditionalTesting={'test to be done'})*

*Example:*

*TnmEnt(TNM='cT3N1M0', Datetime={'01/01/2020'}, AdditionalTesting={'CT Chest, abdomen, Pelvis'})*

*TnmEnt(TNM='T2N1', Datetime={'unknown'}, AdditionalTesting={'unknown'})*

  *f. Grade of tumor*

*For this note, please return the combined Nottingham pathological grade for the main cancer.*

*If additional testing must be done to fully grade the patients tumor, please return this within the same format as described below.*

*If any information for the grade is not clear in the note, please return 'unknown'.*

*Please return as namedtuples separated by newlines in the following format:*

*GradeEnt(Grade='Combined grade', Datetime={'Date conducted'}, AdditionalTesting={'test to be done'})*

*Example:*

*GradeEnt(Grade='3', Datetime={'01/01/2020'}, AdditionalTesting={'unknown'})*

*GradeEnt(Grade='GX', Datetime={'unknown'}, AdditionalTesting={'Number of mitoses'})*

8. Prescribed medications

 *For this note, please return all cancer-directed medications that were prescribed to the patient.*

 *Pair these medications with the date they were prescribed, and the date they were stopped as accurately as possible.*

 *If the medication name has been identified, add the details of the symptom or clinical finding that it was prescribed for.*

 *Also add details about the medication's continuity status among the following options: 'continuing', 'finished', 'discontinued early', or 'unknown'.*

 *Additionally include any problems that were caused due to the medication, and any potential problems that the medication can cause, only if it is mentioned in text.*

 *If any information is not present within the note, please return 'unknown'.*

 *Please return as namedtuples separated by newlines in the following format:*

*PrescribedMedEnt(MedicationName='Medication identified', Begin={'Medication start date or time'}, End={'Medication end date or time'}, Reason={'symptom or clincal finding that the known medication was prescribed for'}, Continuity='continuity status of the medication', ConfirmedAdvEvent={'problems that were certainly caused due to the medication'}, PotentialAdvEvent={'problems that could potentially be caused due to the medication, but did not certainly happen.'})*

*Example:*

*PrescribedMedEnt(MedicationName='Anastrozole', Begin={'01/01/2019'}, End={'01/01/2020'}, Reason={'unknown'}, Continuity='finished', ConfirmedAdvEvent={'swelling'}, PotentialAdvEvent={'unknown'})*

*PrescribedMedEnt(MedicationName='Abraxane', Begin={'01/03/2022'}, End={'unknown'}, Reason={'cancer'}, Continuity='started', ConfirmedAdvEvent={'unknown'}, PotentialAdvEvent={'swelling'})*

*Please do not provide an answer if the MedicationName itself is not available.*
*DO NOT return medications that are planned or may be given in future.*
*Do not skip any fields in the given format.*

9. Future medications

   *For this note, please return all cancer-directed medications that are being considered for future therapy. Please include drug classes if the medication isn't specifically mentioned.*
   *For each, please return whether its considerations is 'planned', or 'hypothetical'.*
   *Also return any potential problems that the medication could cause, only if it is mentioned in text.*
   *Please return as namedtuples separated by newlines in the following format:*
   *FutureMedEnt(MedicationName='Medication identified', Consideration='Classification of consideration', PotentialAdvEvent={'problems that could potentially be caused due to the medication'})*
   *Example:*

*FutureMedEnt(MedicationName='Anastrozole', Consideration='planned', PotentialAdvEvent={'swelling'})*

*FutureMedEnt(MedicationName='Hormone replacement therapy', Consideration='hypothetical', PotentialAdvEvent={'unknown'})*

*DO NOT return medications that were previously prescribed or are currently being prescribed.*

*Do not skip any fields in the given format.*